\documentclass[preprint,12pt,authoryear]{elsarticle}

\usepackage[utf8]{inputenc}
\usepackage{amsmath}
\usepackage{amsfonts}
\usepackage{amssymb}
\usepackage{mathtools}

\usepackage{url}

\usepackage{algorithm}
\usepackage[noend]{algpseudocode}

\usepackage{multirow}
\usepackage[table,xcdraw]{xcolor}

\usepackage{graphicx}
\usepackage{caption}
\usepackage{listings}
\usepackage{verbatim}
\usepackage{wrapfig}
\usepackage{array}
\usepackage{subfigure}

\usepackage{tabularx}

\usepackage{caption}

\usepackage[normalem]{ulem}

\newtheorem{remark}{Remark}

\journal{Engineering Applications of Artificial Intelligence}

\begin{document}

\begin{frontmatter}

\title{Fractal interpolation in the context of prediction accuracy optimization}

\author[label1]{Alexandra Băicoianu}

\author[label1]{Cristina Gabriela Gavrilă}

\author[label1]{Cristina Maria Pacurar\corref{cor1}}

\affiliation[label1]{organization={Faculty of Mathematics and Computer Science, Transilvania University of Brașov},
            addressline={50 Iuliu Maniu Blvd.}, 
            city={Brașov},
            country={Romania}}
\cortext[cor1]{cristina.pacurar@unitbv.ro}

\author[label4]{Victor Dan Pacurar}

\affiliation[label4]{organization={Faculty of Silviculture and Forest Engineering, Transilvania University of Brașov},
            addressline={1 Șirul Beethoven Street}, 
            city={Brașov},
            country={Romania}}

\begin{abstract}
This paper focuses on the hypothesis of optimizing time series predictions using fractal interpolation techniques. In general, the accuracy of machine learning model predictions is closely related to the quality and quantitative aspects of the data used, following the principle of \textit{garbage-in, garbage-out}. In order to quantitatively and qualitatively augment datasets, one of the most prevalent concerns of data scientists is to generate synthetic data, which should follow as closely as possible the actual pattern of the original data. 

This study proposes three different data augmentation strategies based on fractal interpolation, namely the \textit{Closest Hurst Strategy}, \textit{Closest Values Strategy} and \textit{Formula Strategy}. To validate the strategies, we used four public datasets from the literature, as well as a private dataset obtained from meteorological records in the city of Bra\c sov, Romania. The prediction results obtained with the LSTM model using the presented interpolation strategies showed a significant accuracy improvement compared to the raw datasets, thus providing a possible answer to practical problems in the field of remote sensing and sensor sensitivity. Moreover, our methodologies answer some optimization-related open questions for the fractal interpolation step using \textit{Optuna} framework.
\end{abstract}

   
	
	
\end{frontmatter}
 
\section{Introduction}\label{introduction}
	
Developing successful Artificial Intelligence (AI) and machine learning (ML) models requires access to immense amounts of high-quality data, as it is widely acknowledged that the performance of most ML models depends on the quantity and diversity of data. However, collecting the necessary amount of labelled training data can be cost-prohibitive. Thus, developing various strategies to improve the quantity and quality of data is of utmost importance.

Our research focuses on the use of interpolation to enhance the quality of the predictions of ML models. The interpolation technique that we adopt is fractal interpolation which provides interpolants that are not necessarily differentiable functions at every point. Since differentiability implies smoothness and a continuous behavior, interpolating data using functions with this property tends to oversimplify or smooth out some of the irregular and rough patterns that are specific to real-world data, especially at smaller scales. This smoothing effect can lead to a loss of crucial information, making the interpolation less suitable for fitting real-world data. On the other hand, fractal interpolation allows interpolants that can capture the inherent roughness and self-similar structures often found in real-world data. These interpolants can better replicate the complexity and irregularity present in natural phenomena.

On one hand, we develop three different strategies for the preprocessing step of data, which all use fractal interpolation. Our first strategy follows a similar approach to the one used by Raubitzek and Neubauer (\cite{Raubitzek}). However, we managed to solve a series of issues, such as answering the question of optimal choice of the vertical scaling factors involved in fractal interpolation. Moreover, we present a detailed scheme of all the steps involved, which makes our research highly reproducible, and through \textit{Optuna} framework we optimize the prediction model presented. The other two strategies, Closest Values Strategy and Formula Strategy are new approaches  that have not been considered in literature before as far as we know.

On the other hand, besides testing our strategies with an ML model fed with public datasets, we also provide examples using real meteorologic data to put our research in the existing research context. This opens a new gate for researchers in the field to obtain much-needed data either when sensors break, or when finer data are needed, based on data recorded at larger time intervals. The response time of a temperature monitoring sensor is producer-dependent and defines how fast the sensor can adapt to temperature changes in a defined period of time, thus influencing the frequency of data logging.  The sampling rate, which determines the time resolution of data, depends on the sensors’ response time (generally correlated with the price of the device). Developing better interpolation techniques could be very useful for enabling time resolution enhancement, and making it possible, for example, to integrate in the predictive models input data (with high time resolution) subsets obtained from the raw data recorded by sensor-loggers installed in the area for climatological purposes (less expensive devices, with a typical sample rate of 1 hour, but with a much better spatial coverage). This matter highly sustains the utility of the present study, which aims to find new and improved techniques for data interpolation, namely for modifying data time resolution.

	

Generating synthetic data, or data augmentation for time-series data, has been an important research issue for many researchers. Among utilization of data augmentation, we mention augmenting sparse datasets (\cite{Forestier}), generating controllable datasets (\cite{Kang}), moving block bootstrap (\cite{Bergmeir}), a.o. For a comprehensive review on time series augmentation for deep learning see \cite{Wen}. Among applications of data augmentation, we also mention time series classification, (\cite{Fazaw}, \cite{Guennec}, \cite{Iwana}, \cite{Kamycki}) or improving the accuracy of forecasting (\cite{Bandara}, \cite{Lee}, \cite{Raubitzek}). Moreover, it is noticeable that research on time-series data augmentation proved interpolation to be a robust method (\cite{Oh}). 

Classical interpolation methods are often used in prediction machine learning techniques (\cite{Belise},  \cite{Jia}, \cite{Meijering}, \cite{Wu}, \cite{Yadav}). Interpolation techniques have proven to be an essential and effective tool in reconstructing incomplete datasets (\cite{Chai}).

Raubitzek and Neubauer have recently introduced fractal interpolation in data preprocessing for machine learning (\cite{Raubitzek}).  However, our approach addressed several challenges, including determining the optimal selection of vertical scaling factors in fractal interpolation. Moreover, we optimize the presented prediction model using the Optuna framework. Two strategies that we use, namely the Closest Values Strategy and the Formula Strategy, represent novel approaches that have not been used before.
	
Fractal interpolation is a method for generating interpolation points between a given set of data $\Delta = \{(x_i,y_i), i \in \{0,1,2, \dots, N\}\}$,  where $N$ is a natural number. The main difference between fractal interpolation and other types of interpolation techniques is the outlook of the result of interpolation, which is a continuous function that is not differentiable everywhere. Thus, fractal interpolation is more relevant for fitting real-world data. Fractal interpolation has applications in a vast range of areas, such as computer graphics (\cite{Manousopoulos}), image compression (\cite{Drakopulos} and  \cite{May}), reconstruction of satellite images (\cite{Chen}), single-image super-resolution procedure (\cite{Zhang}), reconstruction of fingerprint shape (\cite{Bajahzar}), signal processing (\cite{Navascues} and \cite{Zhai}), reconstruction of epidemic curves (\cite{Pacurar}) and others.

Research combining machine learning and fractal analysis features has been performed before in combination with Support-Vector Machine (see \cite{Ni} that focuses on the enhancement of stock trend prediction accuracy by combining a fractal feature selection method with a support vector machine, demonstrating its superiority compared to five other commonly used feature selection methods, and \cite{Wang} where the stock price indexes are forecasted, offering improved accuracy compared to three other commonly used models) or Time-Delayed Neural Network (see \cite{Yakuma} where there is shown an improved short-term prediction accuracy compared to a back propagation-type forward neural network). 
	
\section{Materials}\label{Materials}

\subsection{Datasets}\label{Datasets}

This section presents the experimental datasets used in the current research study. Their properties sustain the significance of interpolation techniques in the prediction process but also serve as validation for the methodology that we will propose with respect to the quantitative aspect of the data.

\subsubsection{Meteorological Data}\label{Meteorological Data}
	
The data set was provided by the Forest Meteorology-Climatology Laboratory, from the Faculty of Silviculture and Forest Engineering, Transilvania University of Brasov, more precisely it was extracted from the database recorded by the automatic weather station HOBO\textsuperscript{\textregistered}RX3000 (research-grade), deployed at the S\^anpetru Education and Research Base, located about 10 kilometres north-east of Bra\c sov City Centre (45.71$^o$ N, 25.65$^o$ W).

The temperature and relative humidity values were measured and recorded every 10 minutes by an S‐THB smart sensor (Hoboware, produced by Onset Computer Corporation). This device is designed to operate in a range from ‐40$^o$C to 75$^o$C, with an accuracy of $\pm$0.21$^o$C (from 0$^o$ to 50$^o$C, thus in the temperature range of the three autumn months considered in this study) and a resolution of 0.02$^o$C at 25$^o$C.

The sensor response time is 5 minutes (in air moving 1 m/sec), consequently, the time resolution of temperature data measurements (10 minutes) was adequately established. For the data logger programming an important element is the sampling time interval, which depends on the sensor response time (a shorter sampling interval is not acceptable). This issue highly sustains the utility of the present study, which aims to find new, improved techniques for data interpolation, namely for modifying data time resolution. For studying the regional mountain climate, the Forest Meteorology-Climatology Laboratory operates a dense network of temperature and relative humidity data loggers (HOBO Pro v2 Temp/RH logger) deployed in Postavaru Mountains (on different elevations, aspect, wind exposure etc.) with similar accuracy and resolution,
but with a response time of 40 minutes, which forces the sampling interval to be higher, being consequently set at 1 hour (suitable for climatological studies but with the time resolution enhancement useful for other applications).

For this study, the temperature data were formatted with two decimals, as considered adequate for developing and testing the interpolation technique. For meteorological applications, the temperature data resulting from direct measurements should be rounded to one decimal (corresponding to readings at the ordinary meteorological thermometer).

The data set is composed of 13105 temperature entries recorded between 1 September 2021, 0:00:00 and 30 November 2021, 23:50:00. The file is in a \textit{.csv} format with a size of 385 KB.
	
\subsubsection{Additional Public Datasets}\label{Additional Public Datasets}
 
In view of the research by Raubitzek and Neubauer (\cite{Raubitzek}), we consider four additional public datasets: Shampoo sales with 36 data points, (\cite{Kaggle}), Airline passengers with 144 data points, (\cite{Kaggle}), Annual wheat yields in Austria with 57 data points (\cite{faostat}), and Annual maize yields in Austria with 58 data points (\cite{faostat}).

The consistent differences between our study and the research from \cite{Raubitzek} are convincingly outlined by using the same datasets. Moreover,  comparing the methods on the same datasets highlights the progress in this direction of research that our study provides. 
	
\subsection{Prerequisites}\label{Prerequisites}
	
\subsubsection{Fractal Interpolation}\label{Fractal Interpolation}
	
Fractal interpolation was introduced by Barnsley (\cite{Barnsley}) and it has since been intensively studied and applied.

To interpolate the data set $\Delta = \{(x_i,y_i), i \in \{0,1,2, \dots,N\}\}$ consider the equations
\begin{equation}
    \begin{gathered}
        a_i = \dfrac{x_i-x_{i-1}}{x_N-x_0}\\
        c_i = \dfrac{x_Nx_{i-1} - x_0x_i}{x_N-x_0}\\
        d_i = \dfrac{y_i-y_{i-1}}{x_N-x_0} - s_i  \dfrac{y_N-y_0}{x_N-x_0}\\
        e_i = \dfrac{x_iy_{i-1}-x_0y_i}{x_N-x_0} - s_i \dfrac{x_Ny_0 - x_0y_n}{x_N-x_0},
    \end{gathered}
    \label{eqint}
\end{equation}
where $s_i \in (-1,1)$ is called the vertical scaling factor.

Let the family of functions $f_i : [x_0,x_N] \times Y \to  [x_0,x_N] \times Y$ defined as
\begin{equation}
    f_i\binom{x}{y} = \begin{pmatrix}
        a_i & 0\\
        d_i & s_i
    \end{pmatrix}\begin{pmatrix}
        x \\
        y 
    \end{pmatrix}+ \binom{c_i}{e_i},
    \label{f_i}
\end{equation}
for every $i \in \{1,\dots,N\}$.
	
Given a metric space $(X,d)$, the pair $((X,d),(f_i)_{i \in \{ 1},\dots,N\})$ is called an iterated function system (IFS, for short) if:
\begin{itemize}
    \item[i)] $(X,d)$ is a complete space; 
    \item[ii)] the functions $f_i$ are continuous, for every $i \in \{1,\dots,N\}$.
\end{itemize}
The concept of IFS is a notion due to Hutchinson (\cite{Hutchinson}). For an IFS, the fractal operator $F_{\mathcal{S}}:\mathcal{P}(X)\to \mathcal{P}(X)$ is defined as $F_{\mathcal{S}}(K) = \underset{i\in \{1},\dots,N\}{\cup} f_i(K)$, for every $K \in 
\mathcal{P}(X)$, where $\mathcal{P}(X)$ represents the set of all subsets of $X$.

Taking $X=[x_0,x_N]$ with the Euclidean metric and the functions $f_i$ defined in equation (\ref{f_i}), we obtain an IFS.

The fixed point of the fractal operator associated with an iterated function system composed of the functions $(f_i)_{i \in \{1},\dots,N\}$ is an interpolation function for the system of data $\Delta$ called the fractal interpolation function.

We construct the fractal interpolation part based on the above formulas and the code provided by Barnsley in Chapter IV of \emph{Fractals everywhere} (\cite{Barnsley}).

\subsubsection{Optuna Framework}\label{Optuna Framework}
 
\textit{Optuna} is an open-source hyperparameter optimization framework that provides multiple state-of-the-art algorithms for sampling hyperparameters ranging from grid sampling strategies to genetic algorithms approaches (\cite{Optuna1}).

The main steps for using \textit{Optuna} are as follows:
\begin{itemize}
    \item[-] Define an objective function to be optimized.
    \item[-] Create an optimization study  $optuna\_study$, which will determine the best parameters by running several trials. A trial can be defined as a single execution of the objective function.
    \item[-] Use one or multiple \textit{suggest} API function calls for the parameters that are subject to optimization inside a trial.
\end{itemize}
	
The default hyperparameter sampler is TPESampler which implements the Tree-structured Parzen Estimator algorithm (\cite{Optuna2}). The algorithm starts by running the objective function on randomly sampled hyperparameter values from the given domain. After a number of observations, the results are divided into two groups depending on whether they fall below or above a certain quantile of the observed values of the objective function, thus separating the best hyperparameter values from the others. In every iteration, the two groups are updated and a Gaussian Mixture Model (GMM) is fitted to each group, resulting in two densities, $l(x)$ for the best hyperparameters values and $g(x)$ for the remaining ones, where $x$ is the value of the hyperparameter. The algorithm will select the value that maximizes the ratio $\frac{l(x)}{g(x)}$.
	
\section{Method and Procedures}\label{Method and Procedures}
	
\begin{figure}[!htp]
    \centering	
    \includegraphics[width=\textwidth]{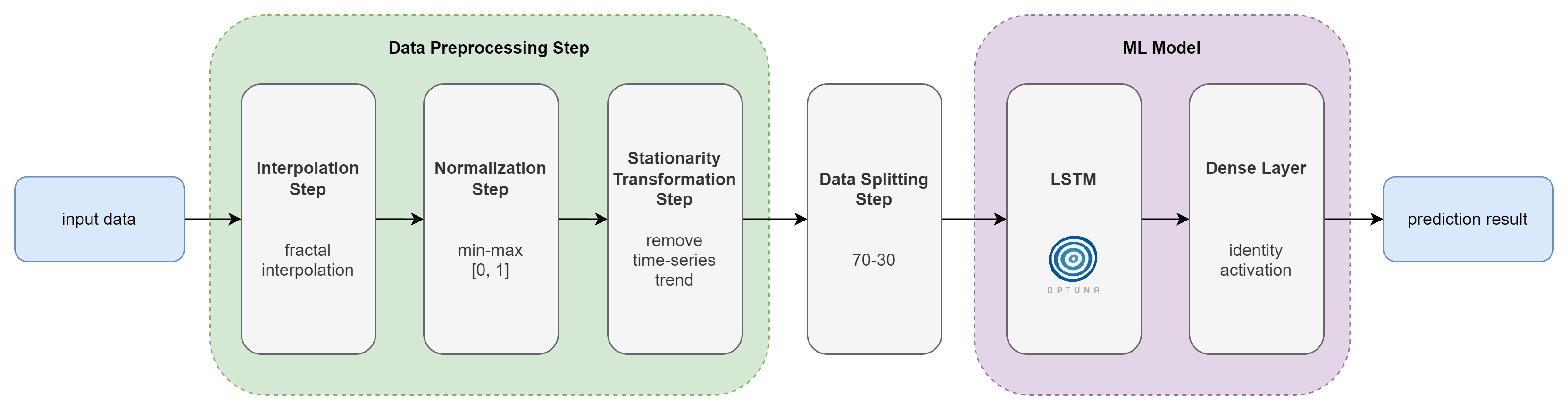}
    \caption{Methodology Outline}
    \label{fig:methodology}
\end{figure}

This section presents the methodology of the present study. We emphasize the steps which are of utmost importance for the final results in Figure \ref{fig:methodology}. Furthermore, in the following sections, we will explore each block included in the diagram and highlight its role in the whole process. 

It is noteworthy that the main contribution of this paper is concentrated on the interpolation step, which is placed in a time series prediction pipeline. The main goal is to evaluate how the data augmentation step, through fractal interpolation, can have an impact on the quality/accuracy of the prediction.

The field of modelling weather and climate is getting increasingly popular, so choosing a suitable learning machine model becomes a challenge. LSTM models are particularly suited for predicting climate change because they can recall and utilize past data to inform future forecasts. Seasonality and patterns in climate data are frequently visible and can last many years. Based on past data, LSTM models may successfully identify these patterns and produce precise forecasts. The capacity of LSTM models to manage missing data is one of their key characteristics. Due to several issues, including sensor malfunctions and data gathering gaps, climate data is frequently unreliable. While LSTM models can make predictions even with insufficient data, traditional statistical methods have difficulty handling missing data.  
 
LSTM models can also be trained to adjust to changing situations. Since climate change is dynamic, conventional statistical models frequently have difficulty adjusting to new patterns and trends. The ability to train LSTM models on an ongoing stream of data, on the other hand, enables them to adjust and produce precise forecasts even as the climate changes.

For all these reasons exposed, in this research, we chose an LSTM model to explore the predictions on the data used.

\subsection{Data Preprocessing Step}\label{Data Preprocessing Step}

Data preprocessing is an essential step in the development of successful ML models. This technique requires some data preparation, including cleaning the data and transforming the data such that their quality is enhanced. Incomplete, inconsistent, or inaccurate data that contain errors or outliers can be eliminated in this preprocessing step. There are numerous preprocessing techniques (for a comprehensive book on data preprocessing (\cite{Garcia}) that produce quality data that lead to high-quality patterns. 

Our preprocessing step includes transformations of data (interpolation, normalization, and stationarity) to obtain the most suitable data for applying ML algorithms. 
	
\subsubsection{Interpolation Step}\label{Interpolation Step}
 
Real-world data are often noisy, with incorrect or missing values. Interpolation is a method of creating new data points within the range of known data points, so it is a technique for filling in missing values. The interpolation should be used where there is a trend observed in the input data and the requirement is to fill the missing value along with the same trend. 

We present the complete and detailed steps required for the interpolation step, applied to the datasets considered. 
	
\textbf{Substep 1: Divide the time series into $m$ subsets of size \emph{sequence\_size}}
	
This step consists of splitting the given sequence into $m$ subsets of length \emph{sequence\_size} so that the last value in subset $i$ is the first value in the subset $i+1$.
	
As regards the way we implement this substep, there are some remarks that are worth mentioning:
	
\begin{remark}
    If the algorithm is used in strict mode, then the dataset is divided into $m$ sequences with equal dimensions; otherwise, the last subset might have a dimension between $3$ and \emph{sequence\_size} - 1 (at least $3$ because $2$ points cannot generate new intermediate points in the interpolation process).
    \label{remark1}
\end{remark} 
	
\begin{remark} 
    At the end of the interpolation, the subsets need to be reunited into a unique list that contains the initial points, chronologically and without repetitions.
    \label{remark2}
\end{remark}
	
\textbf{Substep 2: The proposed interpolation strategies}
	
In this section, we present three different strategies for the steps specific to interpolation, namely \textit{Closest Hurst Strategy (CHS)}, \textit{Closest Values Strategy (CVS)} and \textit{Formula Strategy (FS)}. 
 
We choose different strategies, firstly to obtain validation for the results from \cite{Raubitzek}, and most importantly, to enhance the results and obtain improved techniques. We test our methods for both the public datasets Maize (Annual maize yields in Austria), Shampoo Sales, Airline Passengers, Wheat (Annual wheat yields in Austria), see details in Section \ref{Datasets}, as well as the original data set, Weather described in Subsection \ref{Meteorological Data}. As regards the latter, for all strategies we use the data corresponding to the first week of entries (1 September 2021-8 September 2021) and we chose the data recorded every hour, to better outline the significance of interpolation.

\begin{figure}[!htp]
    \centering							 
    \includegraphics[width=0.7\textwidth]{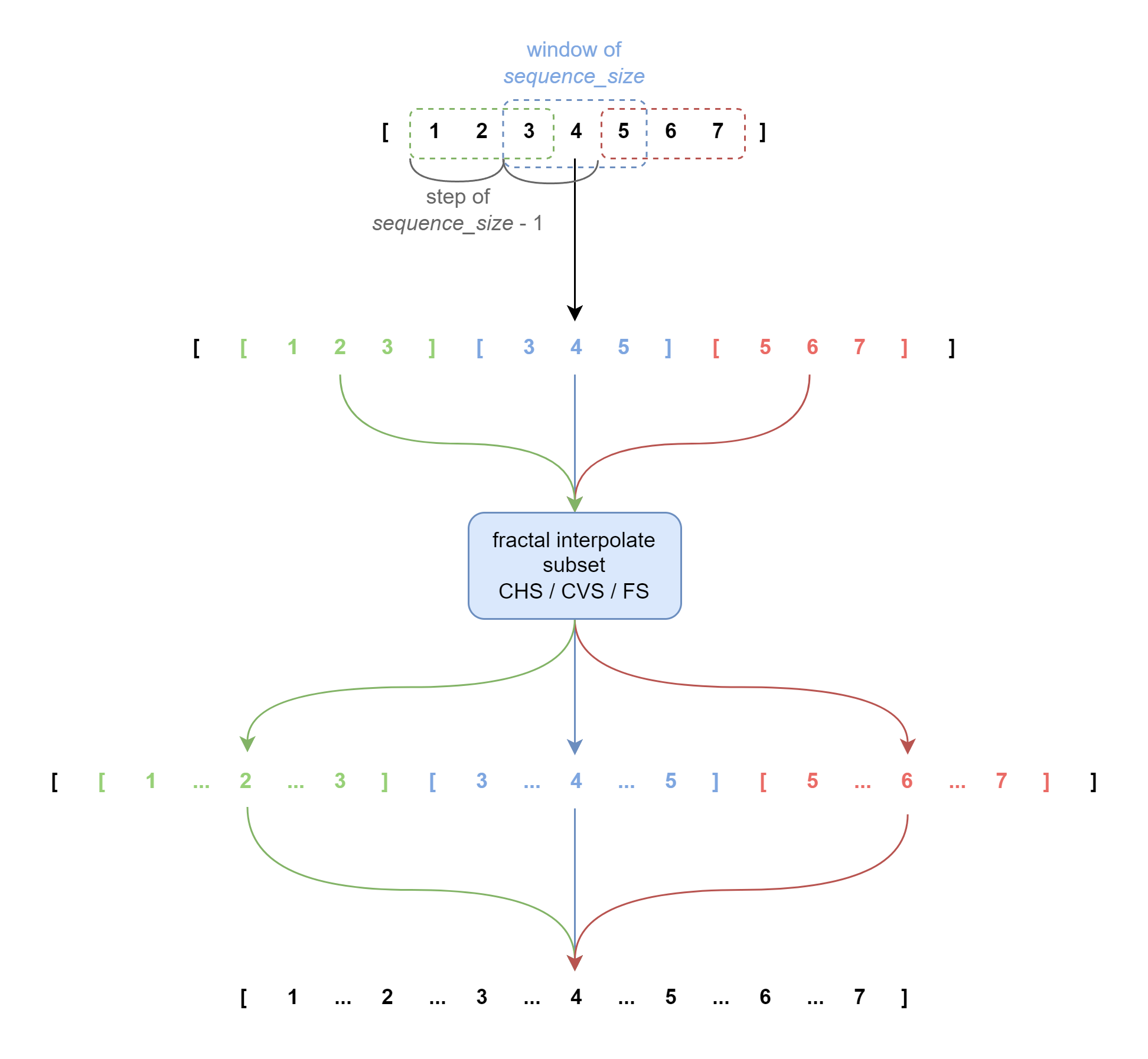}
    \caption{Interpolation flow diagram for $sequence\_size$ 3}
    \label{fig:interpolation_flow}
\end{figure}

Figure \ref{fig:interpolation_flow} describes the general flow of the interpolation step. While the diagram is constructed considering a $sequence\_size$ of $3$, note that a particular $sequence\_size$ was used for implementing the proposed methodologies. Specific details are given in the next sections for each of the defined strategies.

The interpolation step is presented in Algorithm \ref{alg:fractal_interpolation}, and it is applicable for all the next proposed strategies. The description of the parameters for the FRACTAL\_INTERPOLATION procedure are:

\begin{itemize}
    \item[$\circ$] \textit{subset}: subset created from original data as described in \textbf{Substep 1}.
    \item[$\circ$] \textit{$s_i$}: a vector of vertical scaling factors which dictate how jagged (and fractal) will be the aspect of the generated data, in the sense that, as its name states, it scales the points vertically.
    \item[$\circ$] \textit{n\_interpolation}: the number of distinct interpolation points to be generated between every 2 points of the original data
\end{itemize}

\begin{algorithm}
\scriptsize
\begin{algorithmic}[1]
\Procedure{fractal\_interpolation}{$subset,\ s_i,\ n\_interpolation = 17$}
    \State Compute the interpolation factors $a_i, c_i, d_i$ și $e_i$ based on Equation \ref{eqint}
    \State Generate interpolation points based on Equation \ref{f_i} until between every 2 points from the $subset$ there are $n\_interpolation$ distinct interpolation points
\EndProcedure
\end{algorithmic}
\caption{Pseudocode for Fractal Interpolation Computation}
\label{alg:fractal_interpolation}
\end{algorithm}

 \subsubsection*{I. Closest Hurst Strategy (CHS)}\label{I.CHS}
	
The first Strategy is similar to the one employed by \cite{Raubitzek}. We will refer to it as Closest Hurst Strategy (CHS).

For each subset resulting from Interpolation Substep 1  with $sequence\_size$ 10, the Algorithm \ref{alg:CHS} is performed. Note that the parameters have the same signification as previously described.
 
\begin{algorithm}
    \scriptsize
    \begin{algorithmic}[1]
        \Procedure{closest\_hurst\_strategy}{$subset,\ n\_interpolation = 17$}
        \State Compute the $initial\_hurst$, the initial Hurst exponent
        \State Generate $s_i \in [-1, 1]$ a vector with the same value on all positions, representing the constant vertical scaling factor for the current $subset$
        \For{$k \gets 1, 15$}
                \State $interpolated\_subset$ $\gets$ \Call{fractal\_interpolation}{$subset,\ s_i,\ n\_interpolation$}
                \State Compute the $h\_new$, the Hurst exponent for the $interpolated\_subset$
                \If{k = 1}
                    \State $h\_old \gets h\_new$
                    \State $interpolated\_result \gets interpolated\_subset$
                \Else{}
                    \If{abs($h\_new$ - $initial\_hurst$) < abs($h\_old$ - $initial\_hurst$)}
                        \State $h\_old \gets h\_new$
                        \State $interpolated\_result \gets interpolated\_subset$
                    \Else{}
                        \State Generate a new $s_i \in [-1, 1]$
                    \EndIf
                \EndIf
        \EndFor
        \State \textbf{return} $interpolated\_result$
        \EndProcedure
    \end{algorithmic}
    \caption{Pseudocode of Closest Hurst Strategy}
    \label{alg:CHS}
\end{algorithm}

\subsubsection*{Results and Analysis for Closest Hurst Strategy}\label{Results and Analysis for Closest Hurst Strategy}
		
\begin{figure}[!htp]
    \minipage{0.45\textwidth}
        \includegraphics[width=\linewidth]{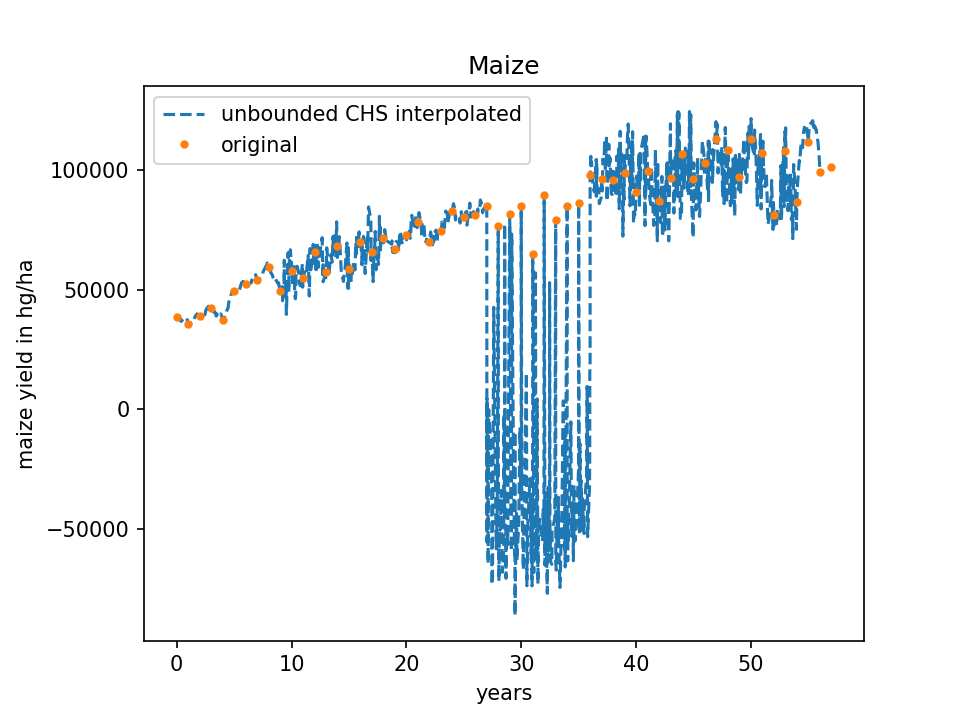}
        \caption{Maize Data set, CHS}
        \label{fig:HurstMaize}
    \endminipage
    \hfill
    \minipage{0.45\textwidth}
        \includegraphics[width=\linewidth]{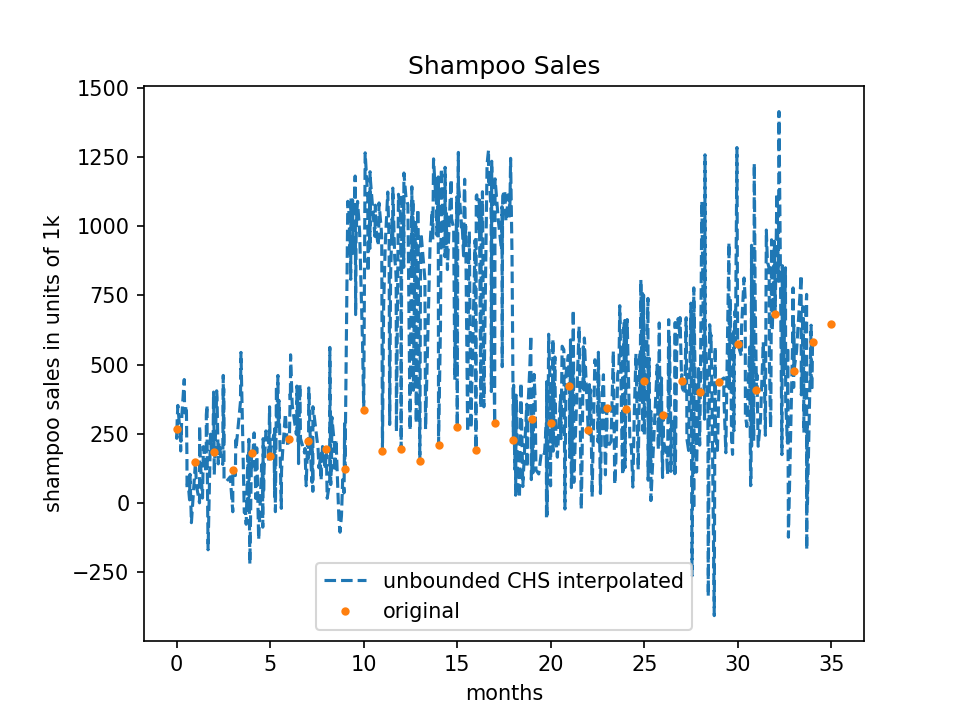}
        \caption{Shampoo Sales Data set, CHS}
        \label{fig:HurstShampoo}
    \endminipage
\end{figure}

\begin{figure}[!htp]
    \minipage{0.45\textwidth}
        \includegraphics[width=\linewidth]{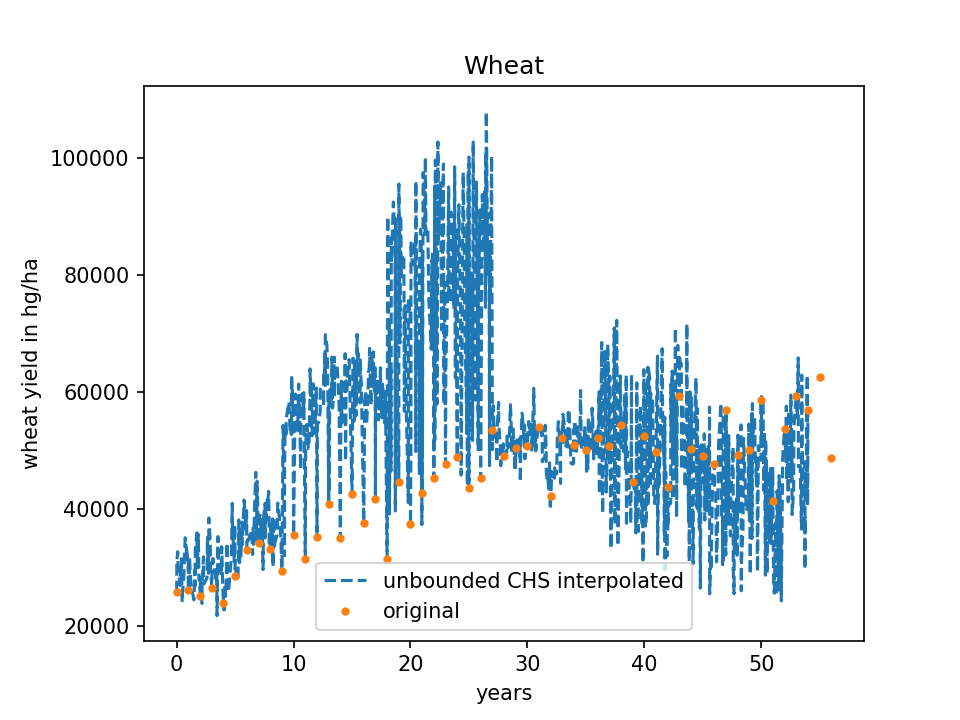}
        \caption{Wheat Data set, CHS}
        \label{fig:HurstWheat}
    \endminipage
    \hfill
    \minipage{0.45\textwidth}
        \includegraphics[width=\linewidth]{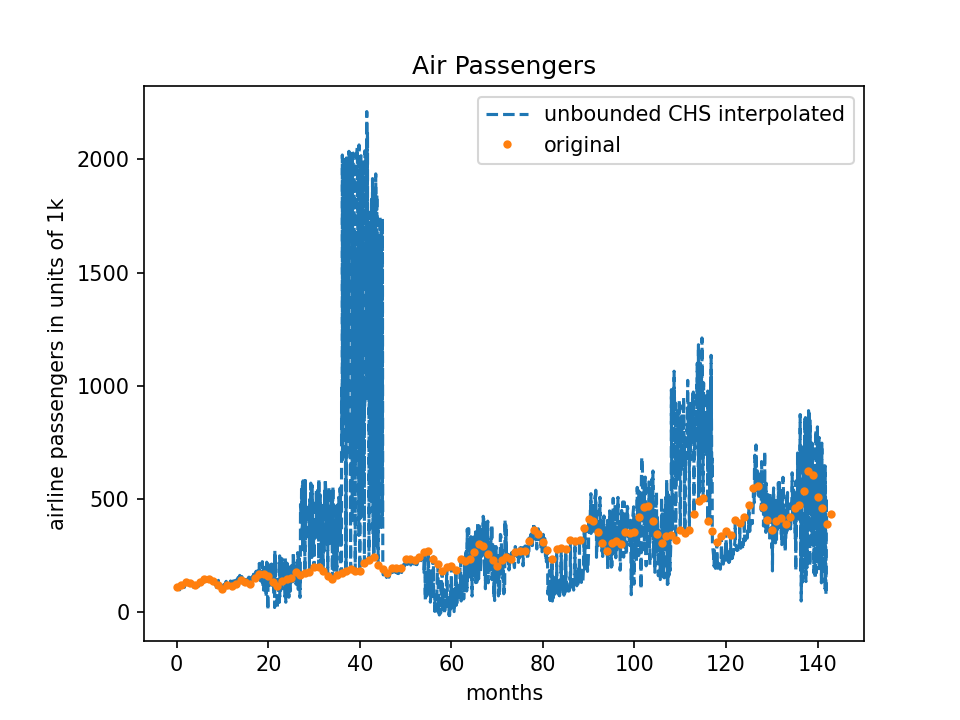}
        \caption{Air Passengers Data set, CHS}
        \label{fig:HurstAir}
    \endminipage
\end{figure}

We present the results obtained for the considered datasets based on CHS. Firstly, we show the outcome for the public datasets, with the parameter $s_i \in [-1,1]$ (Figures \ref{fig:HurstMaize} - \ref{fig:HurstAir}).
	
\begin{figure}[!htp]
    \minipage{0.45\textwidth}
        \includegraphics[width=\linewidth]{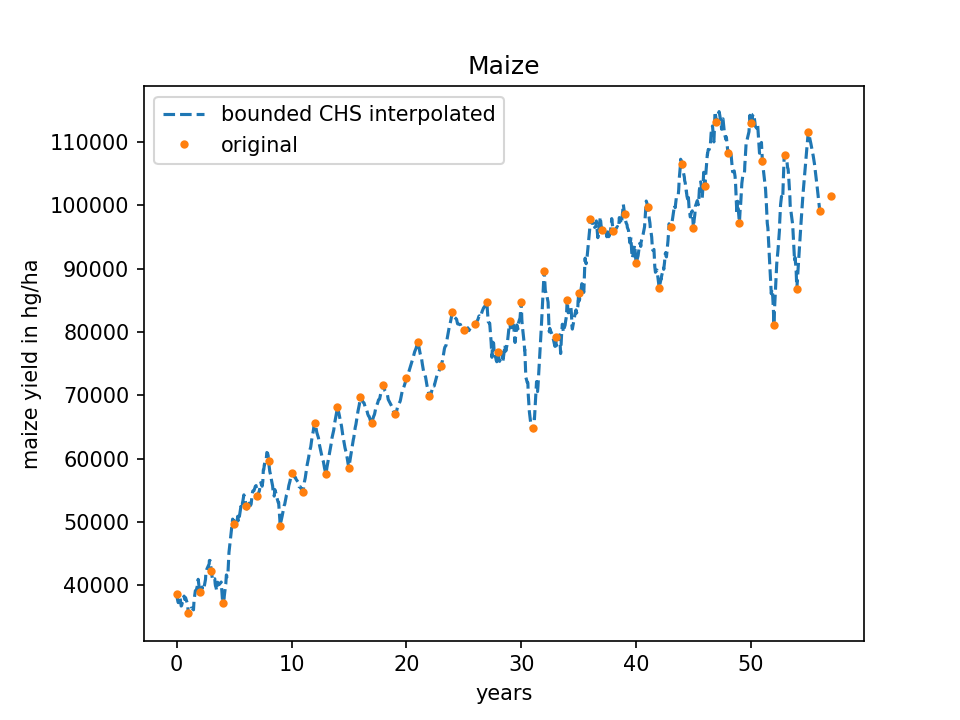}
        \caption{Maize Data set, CHS}
        \label{fig:HurstMaize-1}
    \endminipage
    \hfill
    \minipage{0.45\textwidth}
        \includegraphics[width=\linewidth]{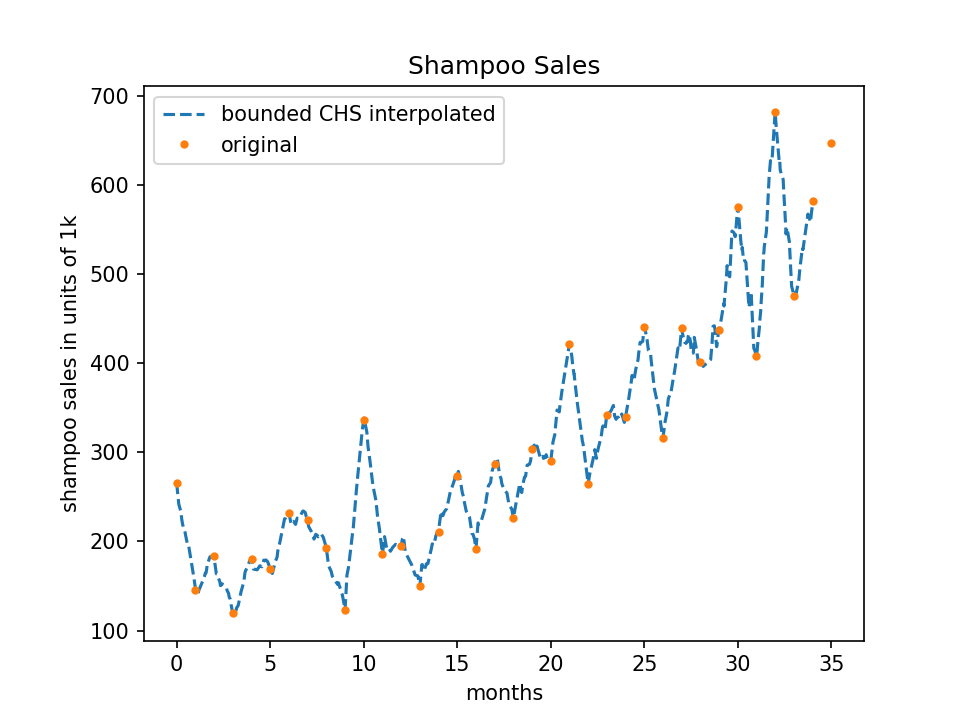}
        \caption{Shampoo Sales Data set, CHS}
        \label{fig:HurstShampoo-1}
    \endminipage
\end{figure}

\begin{figure}[!htp]
    \minipage{0.45\textwidth}
        \includegraphics[width=\linewidth]{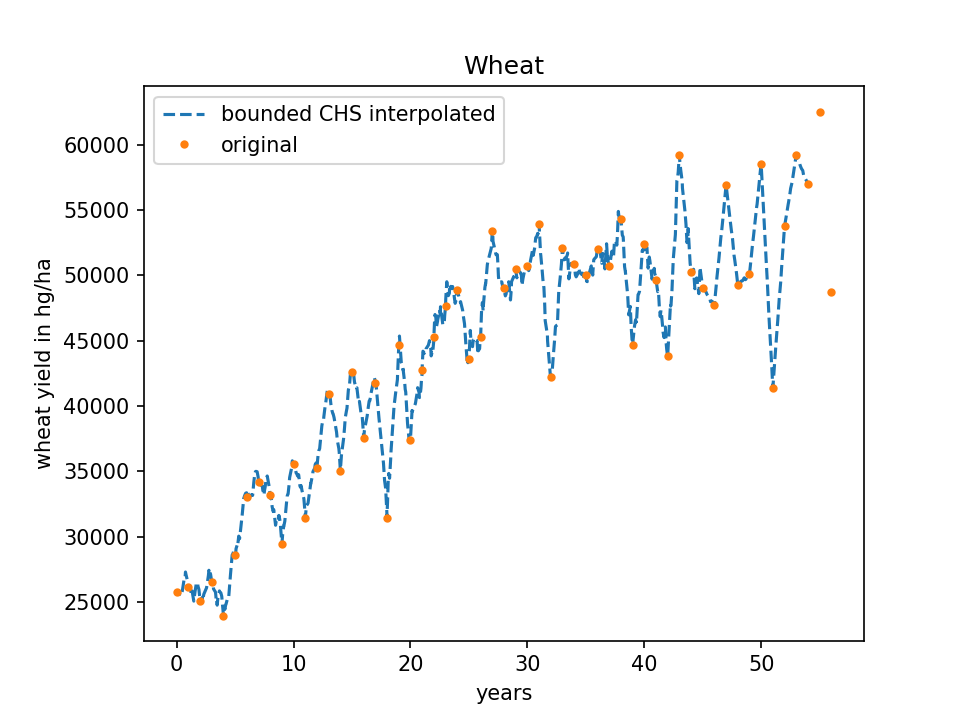}
        \caption{Wheat Data set}
        \label{fig:HurstWheat-1}
    \endminipage
    \hfill
    \minipage{0.45\textwidth}
        \includegraphics[width=\linewidth]{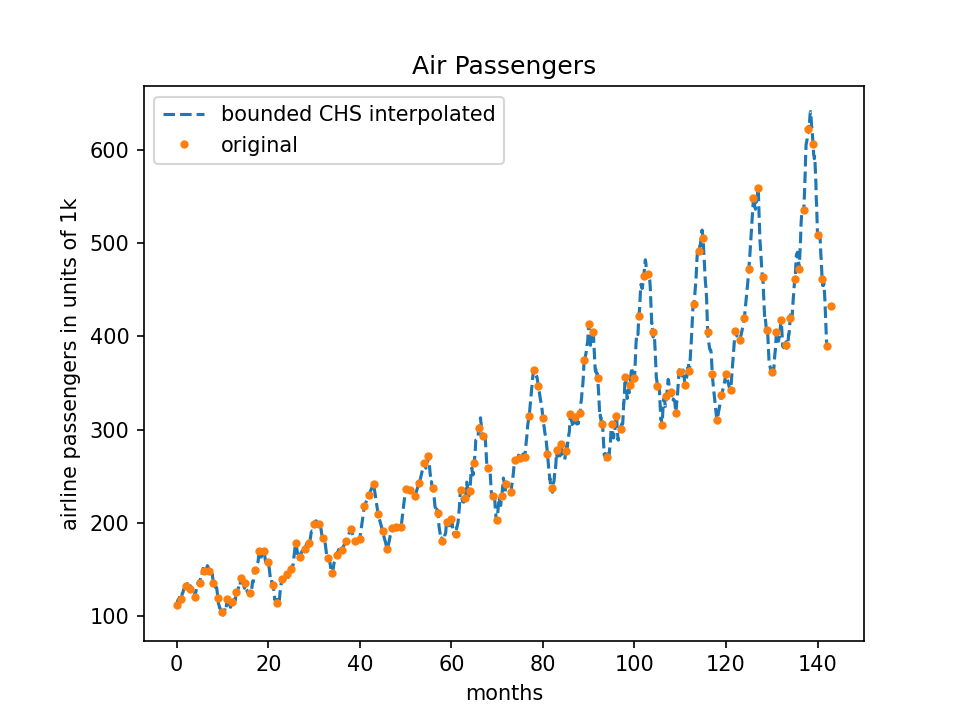}
        \caption{Air Passengers Data set}
        \label{fig:HurstAir-1}
    \endminipage
\end{figure}

However, following tests, we found that the most appropriate vertical scaling factor must be chosen between $s_i \in [0, 0.2]$. In this case, the differences between the points obtained and the initial points are limited, and the initial outlook of the graphic defined by the initial points is conserved. We show in Figures \ref{fig:HurstMaize-1} - \ref{fig:HurstAir-1} the results obtained in this case.
	
\begin{figure}[!htp]
    \minipage{0.45\textwidth}
        \includegraphics[width=\linewidth]{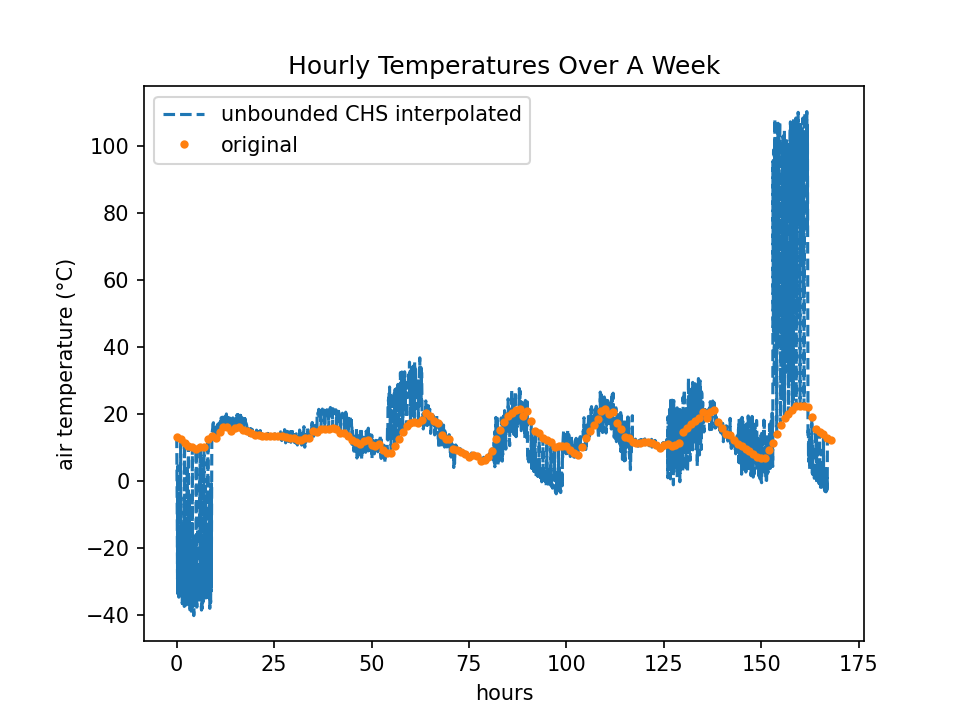}
        \caption{Weather Data set with $s_i \in [-1,1]$}
        \label{fig:HurstWeather}
    \endminipage
    \hfill
    \minipage{0.45\textwidth}
        \includegraphics[width=\linewidth]{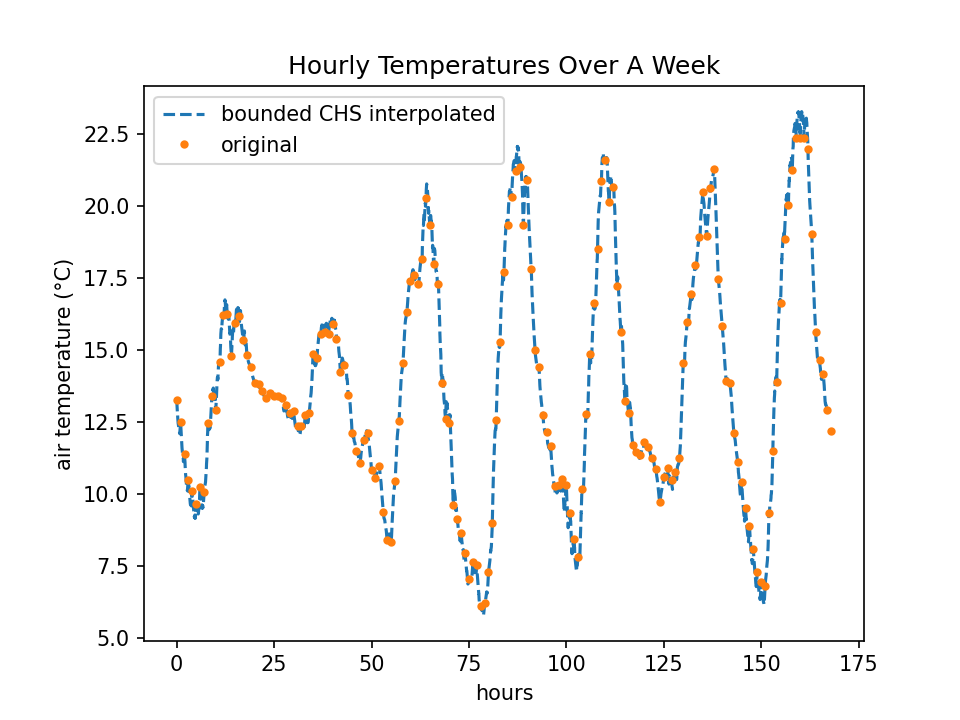}
        \caption{Weather Data set with $s_i \in [0,0.2]$}
        \label{fig:HurstWeather-1}
    \endminipage
\end{figure}

For our private data set, Weather, we obtain the results presented in Figures \ref{fig:HurstWeather} and \ref{fig:HurstWeather-1}.
	
Although the stop condition ensures that the Hurst exponent for the interpolated data is close enough to the initial Hurst value, this does not guarantee the persistence of other properties of the data. This motivates us to define new strategies that ensure the preservation of certain properties of the data through interpolation.
	
\subsubsection*{II. Optimized Procedure - Closest Values Strategy (CVS)}\label{II.CVS}
	
For this type of strategy, we propose the \textit{Optuna} framework, described in Section \ref{Optuna Framework}. 
	
Thus, in the current CVS strategy, for each data subset obtained using $sequence\_size$ 10, the steps from Algorithm \ref{alg:CVS} are performed. Observe that Algorithm \ref{alg:CVS} defines two procedures, the first one is the objective function that \textit{Optuna} will minimize over the course of 15 trials, having the new parameter $optuna\_trial$, and the second one is the main implementation of the strategy. Additionally, procedure LINEAR\_INTERPOLATION constructs a linear interpolation of the $subset$ considering each interpolation point generated by the FRACTAL\_INTERPOLATION procedure, and RMSE represents the Root Mean Square Error.
	
\begin{algorithm}
    \begin{algorithmic}[1]
        \scriptsize
        \Procedure{closest\_values\_strategy\_objective}{$optuna\_trial,\ subset,\ n\_interpolation = 17$}
            \State Generate $s_i \in [-1, 1]$, a vector with the same value on all positions, representing the constant vertical scaling factor for the current $subset$ in the current $optuna\_trial$ using $suggest$ API
            \State $interpolated\_subset$ $\gets$ \Call{fractal\_interpolation}{$subset,\ s_i,\ n\_interpolation$}
            \State $linear\_interpolated\_subset$ $\gets$ \Call{linear\_interpolation}{$subset$}
            \State \textbf{return} \Call{RMSE}{$interpolated\_subset,\ linear\_interpolated\_subset$}
        \EndProcedure
        \Comment{}
        \Procedure{closest\_values\_strategy}{$subset,\ n\_interpolation = 17$}
            \State Create $optuna\_study$, a study with direction 'minimize', the objective function $\Call{closest\_values\_strategy\_objective}$ and 15 trials
            \State $s_i \gets$ best trial parameter of $optuna\_study$
            \State \textbf{return} \Call{fractal\_interpolation}{$subset,\ s_i,\ n\_interpolation$}
        \EndProcedure
    \end{algorithmic}
    \caption{Pseudocode of Closest Values Strategy}
    \label{alg:CVS}
\end{algorithm}

\subsubsection*{Results and Analysis for Closest Values Strategy}\label{Results and Analysis for Closest Values Strategy}
	
\begin{figure}[!htp]
    \minipage{0.45\textwidth}
        \includegraphics[width=\linewidth]{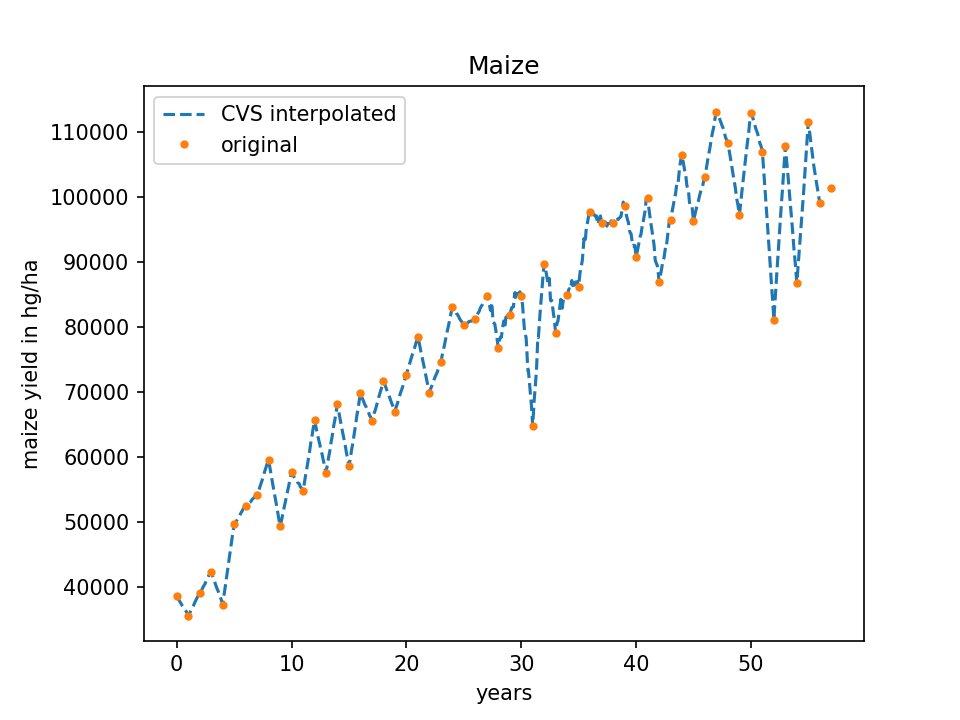}
        \caption{Maize Data set, CVS}
        \label{fig:OptunaMaize}
    \endminipage
    \hfill
    \minipage{0.45\textwidth}
        \includegraphics[width=\linewidth]{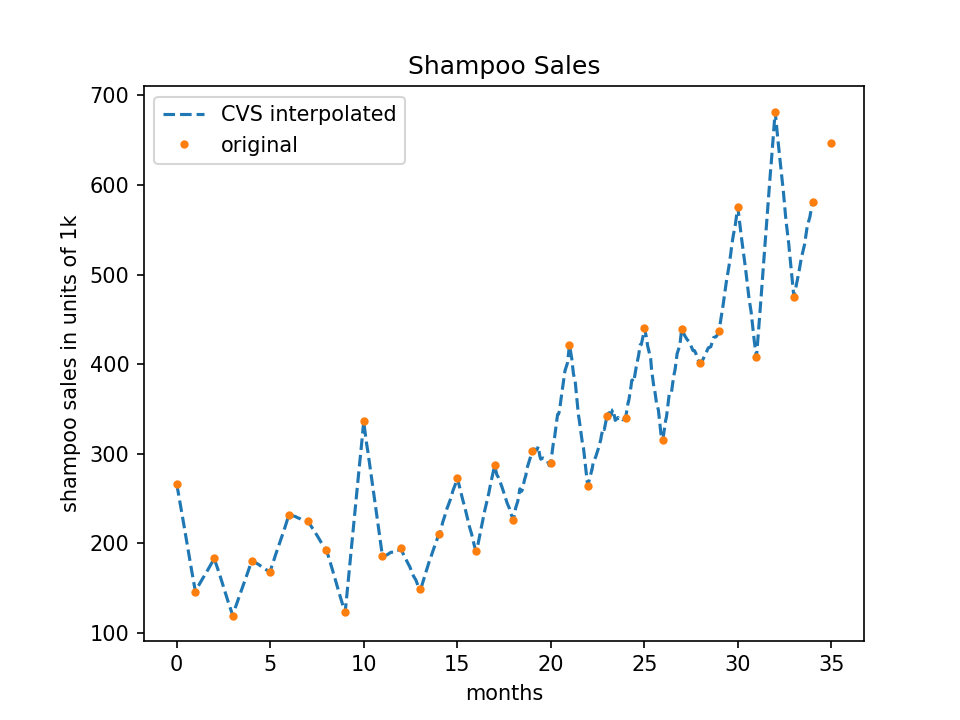}
        \caption{Shampoo Sales Data set, CVS}
        \label{fig:OptunaShampoo}
    \endminipage
\end{figure}

\begin{figure}[!htp]
    \minipage{0.45\textwidth}
        \includegraphics[width=\linewidth]{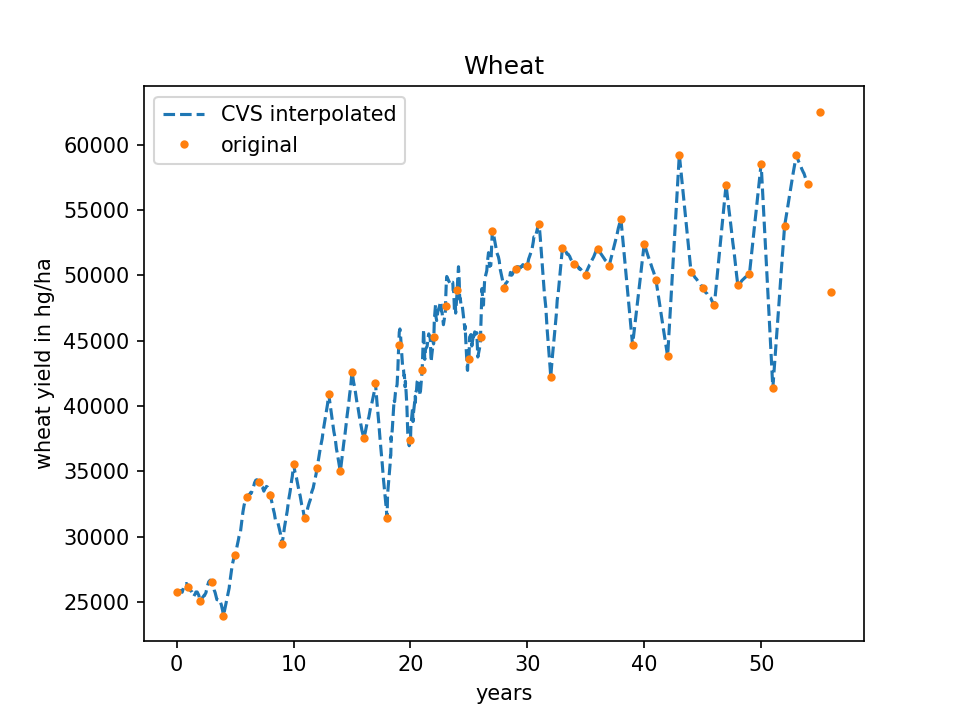}
        \caption{Wheat Data set, CVS}
        \label{fig:OptunaWheat}
    \endminipage
    \hfill
    \minipage{0.45\textwidth}
        \includegraphics[width=\linewidth]{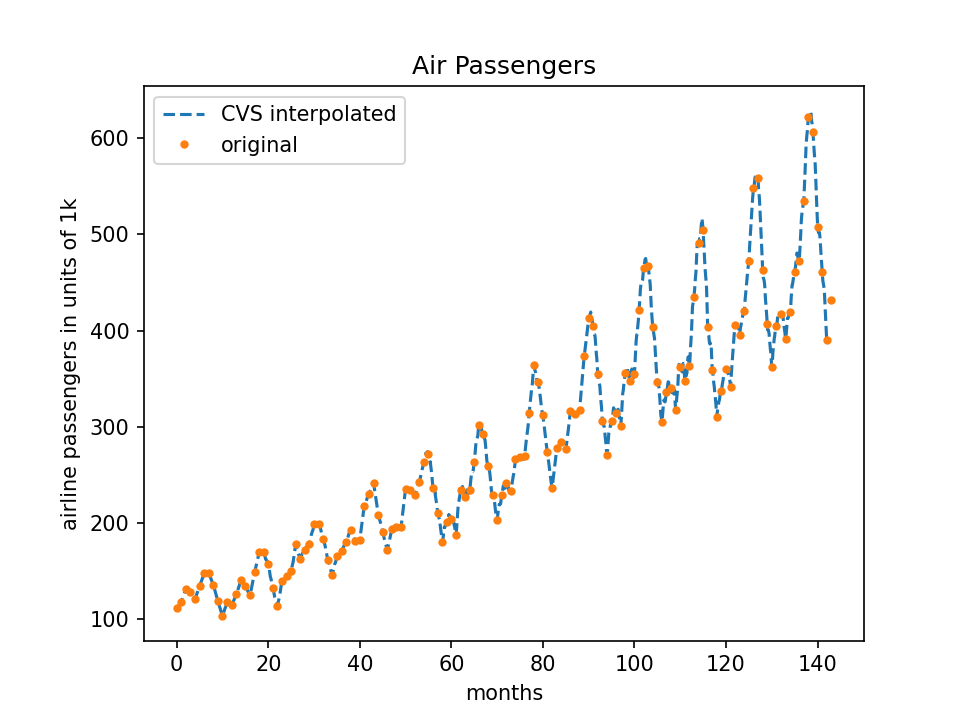}
        \caption{Air Passengers Data set, CVS}
        \label{fig:OptunaAir}
    \endminipage
\end{figure}

\begin{figure}[!htp]
    \centering							 
    \includegraphics[width=0.5\textwidth]{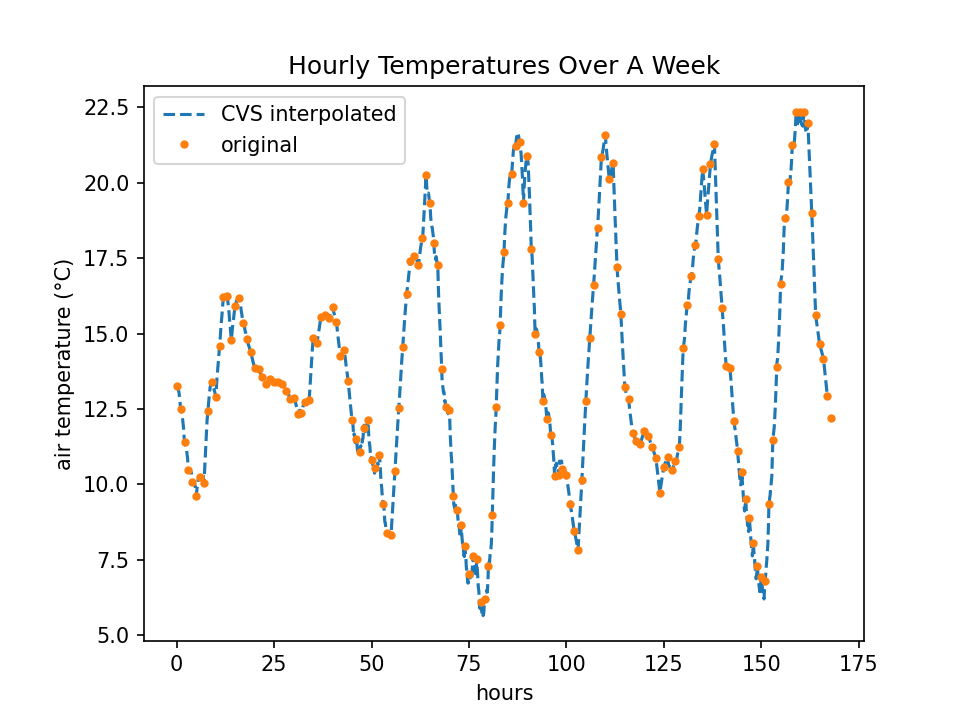}
    \caption{Weather Data set, CVS}
    \label{fig:OptunaWeather}
\end{figure}

In Figures \ref{fig:OptunaMaize} - \ref{fig:OptunaWeather} there are shown the results of the proposed strategy for all five datasets based on CVS. The parameter $s_i$ was optimized by \textit{Optuna} with a possible range set in the interval $[-1,1]$.
	
\begin{figure}[!htp]
    \centering							 
    \includegraphics[width=0.5\textwidth]{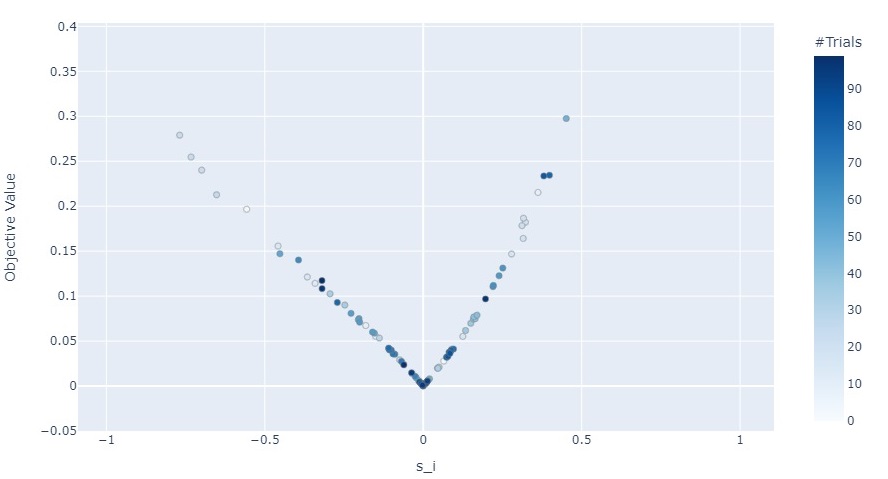}
    \caption{Evolution of parameter $s_i$ with \textit{Optuna}}
    \label{fig:Optuna}
\end{figure}

It is noticeable that the graphics obtained using CVS resemble the results from CHS with $s_i \in [0,0.2]$. This is because the RMSE is minimum for the parameter $s_i$ close to the interval $[0,0.2]$. This can be observed in Figure \ref{fig:Optuna} where the evolution of the parameter $s_i$ is presented with respect to the objective function described in the CVS context. This result validates our choice of the parameter $s_i$ in the case of CHS.
	
\subsubsection*{III. Optimized Strategy - Formula Strategy (FS)}\label{III.FS}
	
For this method, we shall follow a different direction, that is to use a formula to optimize the parameter $s_i$. Although there is no way to determine a general optimal vertical scaling factor, there are various approaches to optimize this parameter. We follow the ideas from \cite{Mazel}, \cite{Manousopoulos1} and \cite{Gowrisankar}. 
	
For the data set $\Delta = \{(x_i,y_i), i \in \{0,1,2, \dots,N\}\}$, we choose the parameter $s_i$ as follows:

\begin{equation}
    s_i = \dfrac{y_i-y_{i-1}}{\sqrt{(y_N-y_0)^2+(y_i-y_{i-1})^2}},
    \label{si}
\end{equation}	
for each $i \in \{1,2, \dots,N\}$.
	
However, since the denominator in equation (\ref{si}) becomes closer to $0$ when the first and the last data in the subset are too close (the line determined by the start point and ending point of the subset is parallel to the $Ox$ axis), then $s_i$ becomes irrelevantly big, inducing an unwanted variation in the data. Thus, we avoid this by optimizing the current strategy. The optimization is achieved by modifying the $sequence\_size$ parameter such that the difference between the two ends does not tend to zero.
	
To exemplify the dependence of the interpolated data on the $sequence\_size$ parameter chosen, let us consider a trial data set 

$$\Gamma = \{(1,10), (2,14), (3,19), (4,26), (5,35), $$
$$(6, 46), (7,35), (8,26), (9,19), (10,14), (11,10)\}.$$ 

\begin{figure}[!htp]
    \minipage{0.45\textwidth}
        \includegraphics[width=\linewidth]{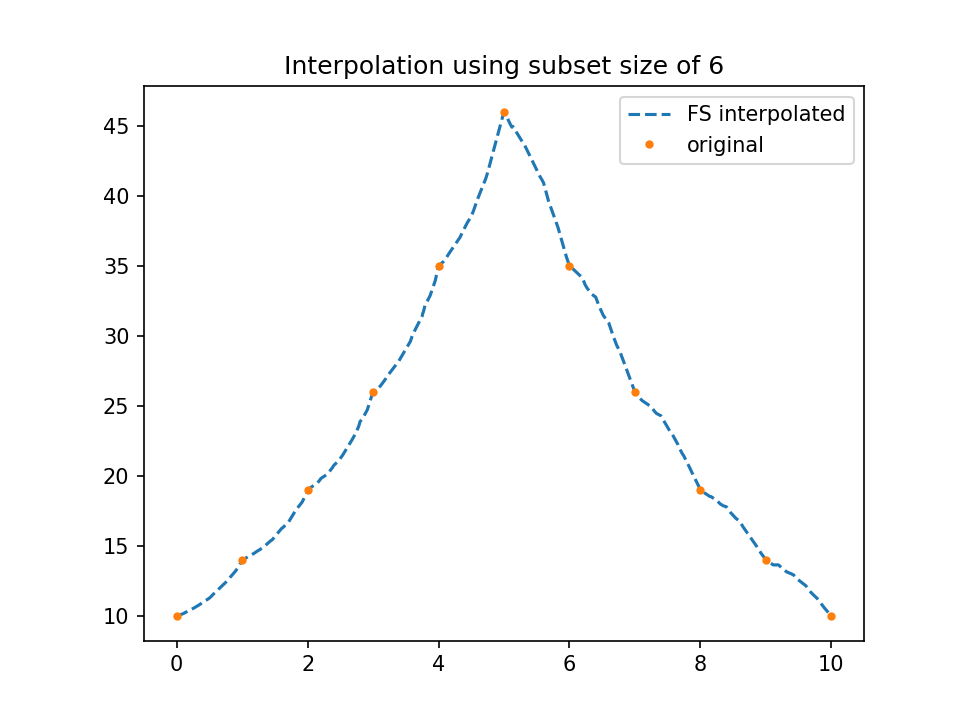}
        \caption{Interpolation of $\Gamma$ with $sequence\_size$ 6}
        \label{fig:Data-l6}
    \endminipage
    \hfill
    \minipage{0.45\textwidth}
        \includegraphics[width=\linewidth]{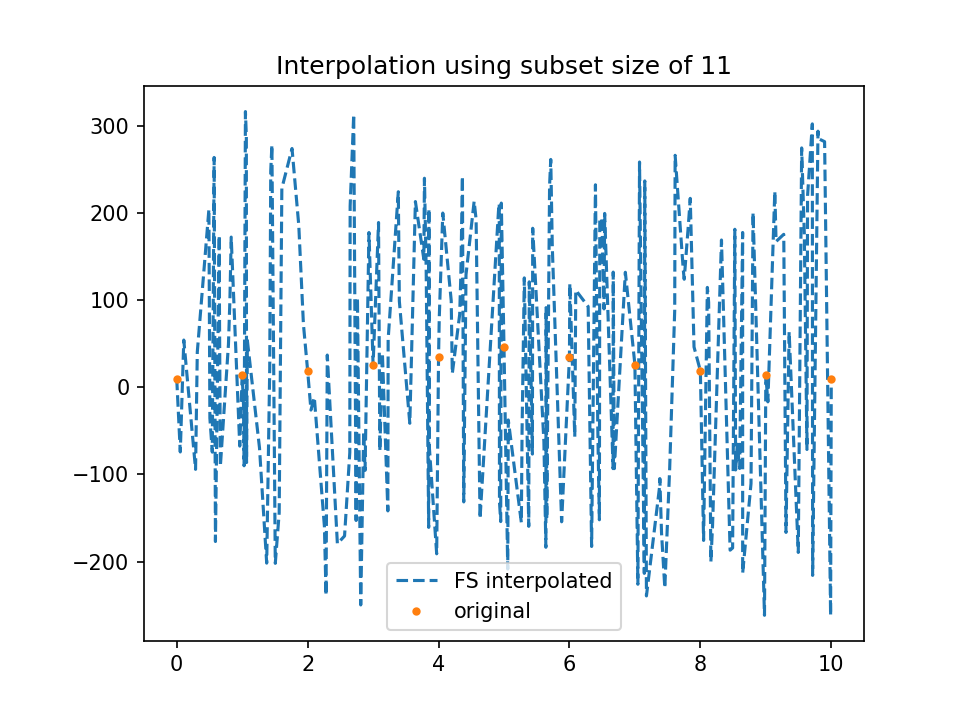}
        \caption{Interpolation of $\Gamma$ with $sequence\_size$ 11}
        \label{fig:Data-l11}
    \endminipage
\end{figure}

In Figures \ref{fig:Data-l6} and \ref{fig:Data-l11}, there are presented the results of interpolation with FS for two different values for $sequence\_size$, $6$ and $11$ respectively.
	
Therefore, in the context of FS, the initial step is determining the optimal  value of the $sequence\_size$. For this, we used the procedures defined in Algorithm \ref{alg:FS_length}. Note that the optimization is computed for the entire dataset using \textit{Optuna} with 50 trials, as opposed to previous optimization strategies where the optimization was done at the subset level.

To determine the optimal values for $sequence\_size$, we consider the search interval $[4, \text{length($dataset$)} - 3]$ and allow the fractal interpolation algorithm to work in the non-strict mode, see Remark \ref{remark1}, to minimize data loss. 

\begin{algorithm}
    \begin{algorithmic}[1]
    \scriptsize
        \Procedure{optimize\_subset\_length\_objective}{$optuna\_trial,\ dataset,\ n\_interpolation = 17$}
            \State Generate subset length $sequence\_size \in [4, \text{length($dataset$)} - 3]$ in the current $optuna\_trial$ using $suggest$ API
            \State Split $dataset$ into $subsets$ of length $sequence\_size$ as described in \textbf{Substep 1} from Section \ref{Interpolation Step}
            \State $total\_RMSE \gets 0$
            \For{$\textbf{each}\ subset\ \textbf{in}\ subsets$}
                \State $interpolated\_subset$ $\gets$ \Call{formula\_strategy}{$subset,\ n\_interpolation$}
                \State $linear\_interpolated\_subset$ $\gets$ \Call{linear\_interpolation}{$subset$}
                \State $total\_RMSE \gets total\_RMSE + \Call{RMSE}{interpolated\_subset,\ linear\_interpolated\_subset}$
            \EndFor
            \State \textbf{return} $total\_RMSE$
        \EndProcedure
        
        \Comment{}
        
        \Procedure{optimize\_subset\_length}{$dataset,\ n\_interpolation = 17$}
            \State Create $optuna\_study$, a study with direction 'minimize', the objective function $\Call{optimize\_subset\_length\_objective}$ and 50 trials
            \State $sequence\_size \gets$ best trial parameter of $optuna\_study$
            \State \textbf{return} $sequence\_size$
        \EndProcedure
    
    \end{algorithmic}
    \caption{Pseudocode of $sequence\_size$ optimization}
    \label{alg:FS_length}
\end{algorithm}

With regards to FS, for each subset of data, the procedure from Algorithm \ref{alg:FS} is executed.
	
\begin{algorithm}
    \begin{algorithmic}[1]
        \scriptsize
        \Procedure{formula\_strategy}{$subset,\ n\_interpolation = 17$}
            \State Compute $s_i$ based on Equation \ref{si}
            \State \textbf{return} \Call{fractal\_interpolation}{$subset,\ s_i,\ n\_interpolation$}
        \EndProcedure
    \end{algorithmic}
    \caption{Pseudocode of Formula Strategy}
    \label{alg:FS}
\end{algorithm}
        
Let us note that for the procedure FORMULA\_STRATEGY, $s_i$ is not constant for the entire subset (as is the case for CHS and CVS), but it varies according to each interval defined by two consecutive points in the subset. 
	
One of the main advantages of this strategy is that once the optimal dimension of the subset is found,  the repetitive process required to execute the optimization routine for the previous two strategies is no longer required.
	
The optimal value of $sequence\_size$ for each data set is presented in Table \ref{length}.
	
\begin{table}[!htp]
    \centering
    \begin{tabular}{|c|c|}
        \hline
        Data set & $sequence\_size$\\
        \hline
        Maize & 29 \\
        \hline
        Shampoo Sales & 10 \\
        \hline
        Wheat & 54 \\
        \hline
        Air Passengers & 141 \\
        \hline
        Weather & 6 \\
        \hline
    \end{tabular}
    \caption{Optimal $sequence\_size$ values for FS interpolation}
    \label{length}
    \vspace{-0.75cm}
\end{table}
	
\subsubsection*{Results and Analysis for Formula Strategy}\label{Results and Analysis for Formula Strategy}
		
\begin{figure}[!htp]
    \minipage{0.45\textwidth}
        \includegraphics[width=\linewidth]{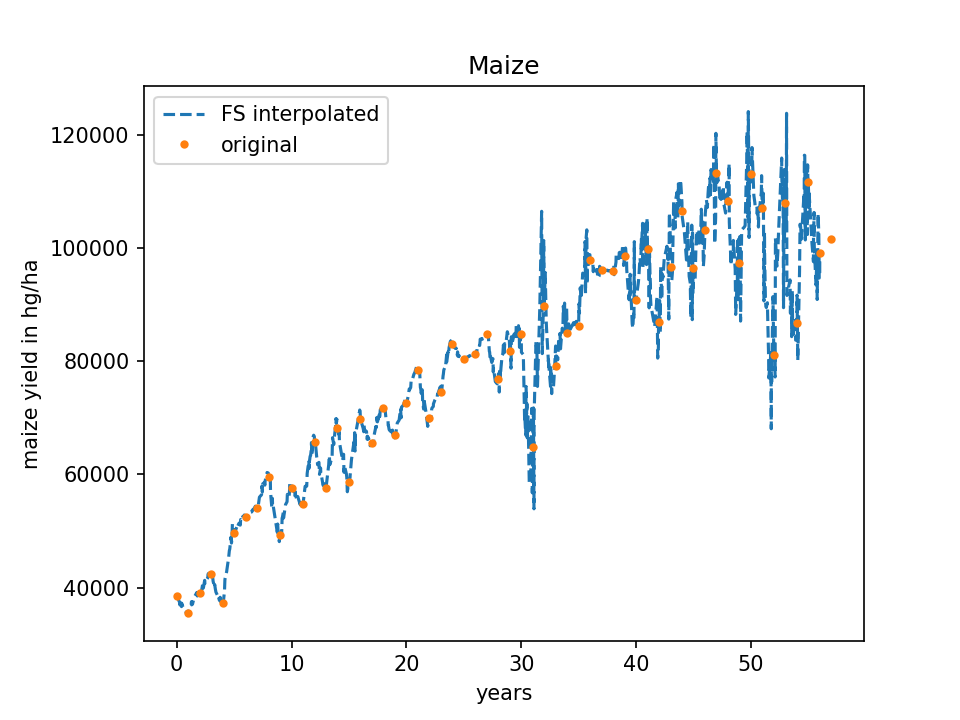}\vspace{-0.5cm}
        \caption{Maize Data set, FS}
        \label{fig:FormulaMaize2}
    \endminipage
    \hfill
    \minipage{0.45\textwidth}
        \includegraphics[width=\linewidth]{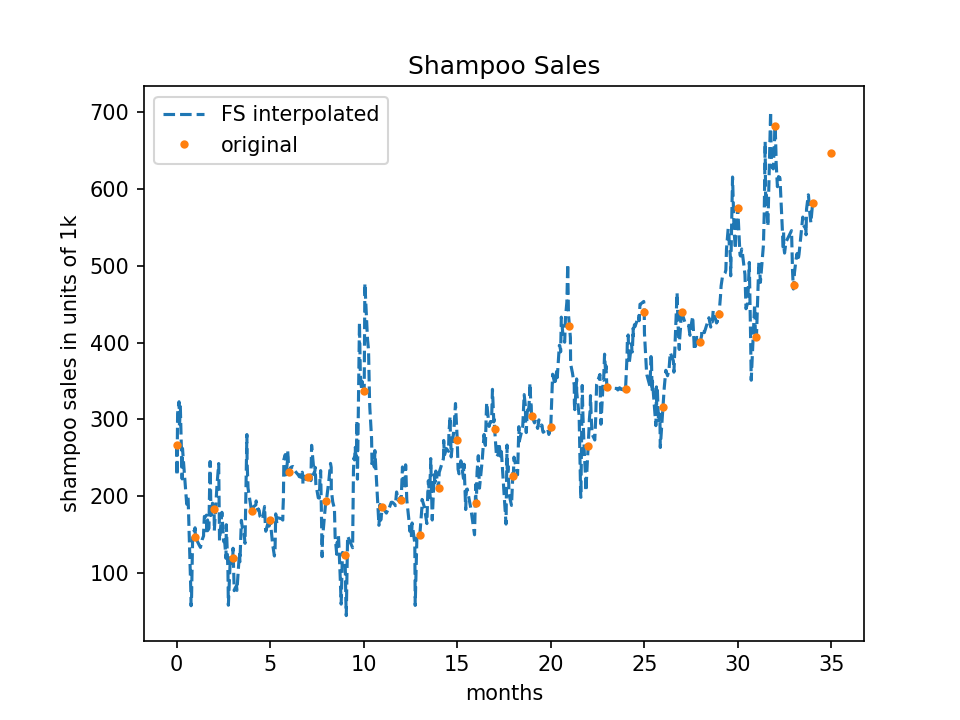}\vspace{-0.5cm}
        \caption{Shampoo Sales Data set, FS}
        \label{fig:FormulaShampoo2}
    \endminipage
\end{figure}

\begin{figure}[!htp]
    \minipage{0.45\textwidth}
        \includegraphics[width=\linewidth]{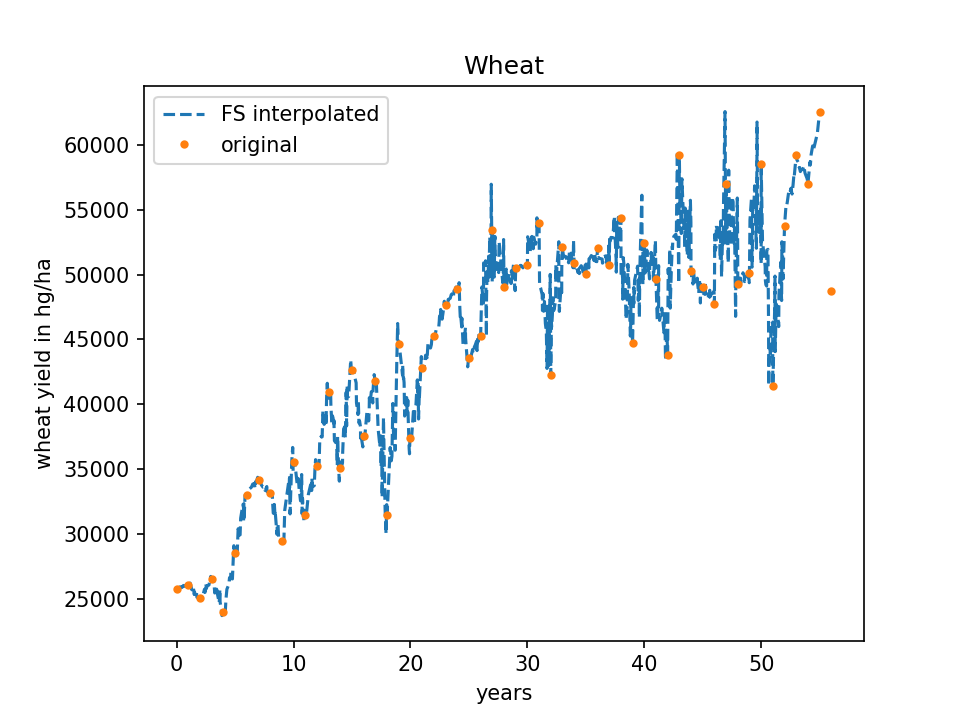}
        \caption{Wheat Data set, FS}
        \label{fig:FormulaWheat2}
    \endminipage
    \hfill
    \minipage{0.45\textwidth}
        \includegraphics[width=\linewidth]{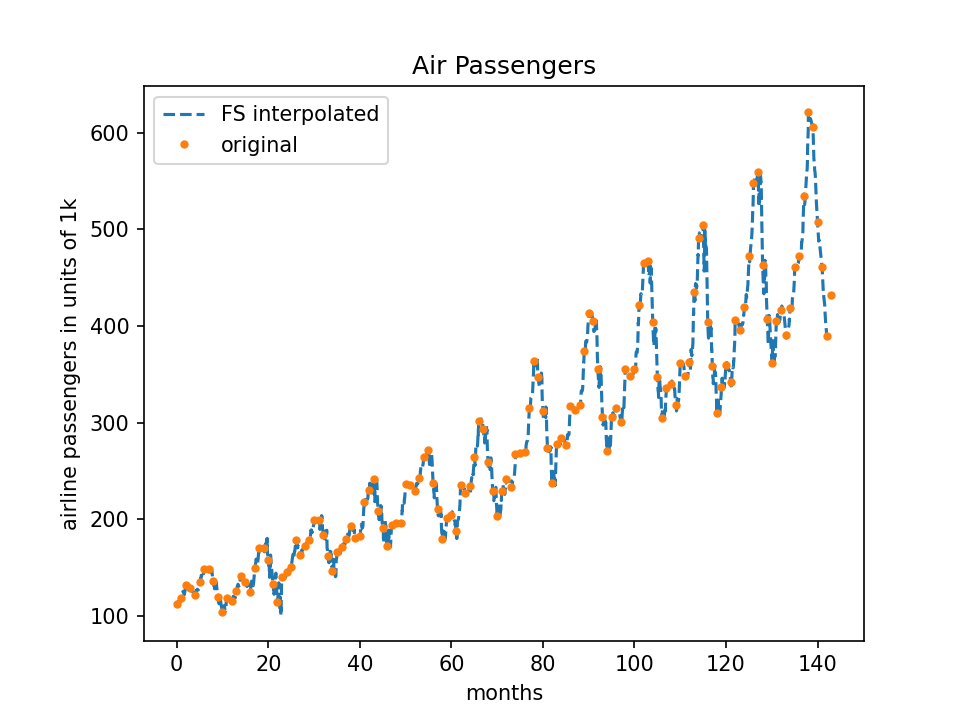}
        \caption{Air Passengers Data set, FS}
        \label{fig:FormulaAir2}
    \endminipage
\end{figure}

\begin{figure}[!htp]
    \centering							 
    \includegraphics[width=0.5\textwidth]{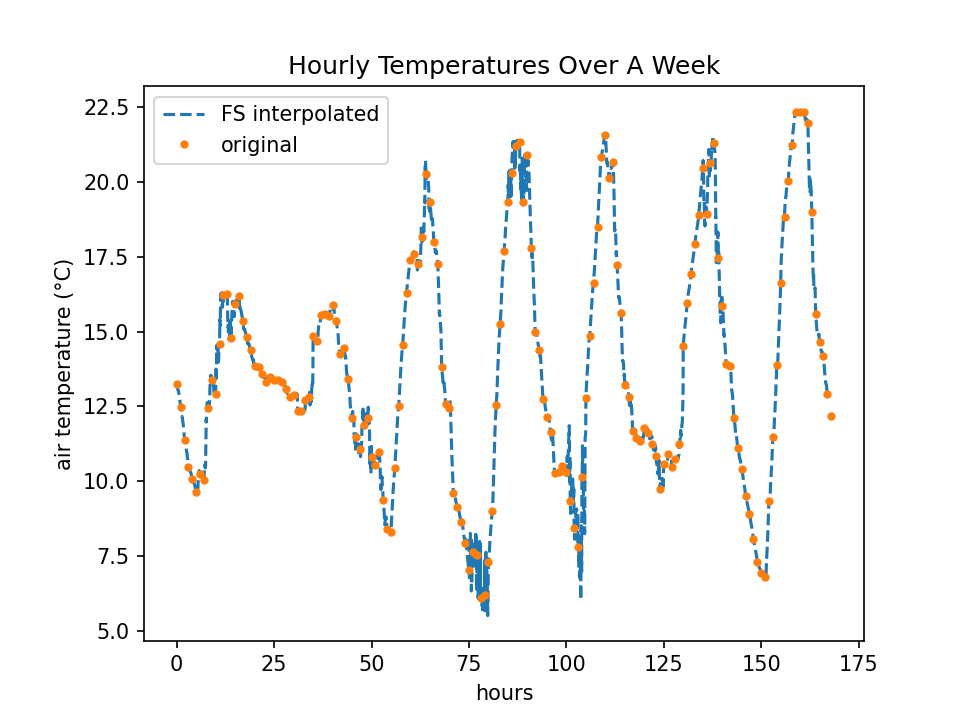}\vspace{-0.25cm}
    \caption{Weather Data set, FS}
    \label{fig:FormulaWeather2}
\end{figure}

Applying formula (\ref{si}) using the optimal $sequence\_size$ values from Table \ref{length}, we obtain the interpolation results presented in Figures \ref{fig:FormulaMaize2} - \ref{fig:FormulaWeather2}.
	
\subsubsection*{Comparison of methods and Results}\label{Comparison of methods and Results}
	
\begin{figure}[!htp]
    \minipage{0.45\textwidth}
        \includegraphics[width=\linewidth]{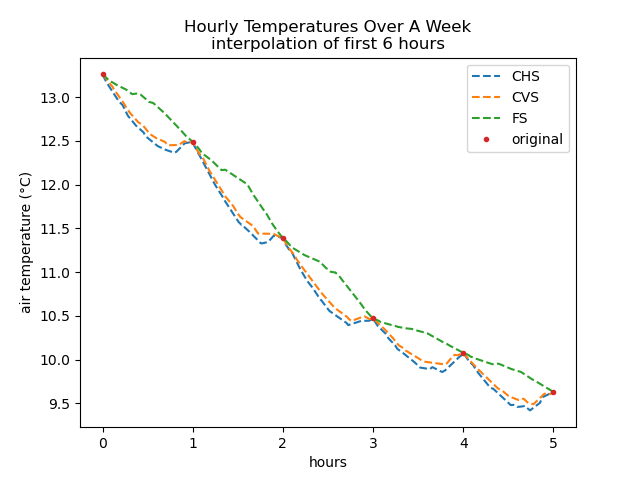}
        \caption{First 6 points}
        \label{fig:Comparison1}
    \endminipage
    \hfill
    \minipage{0.45\textwidth}
        \includegraphics[width=\linewidth]{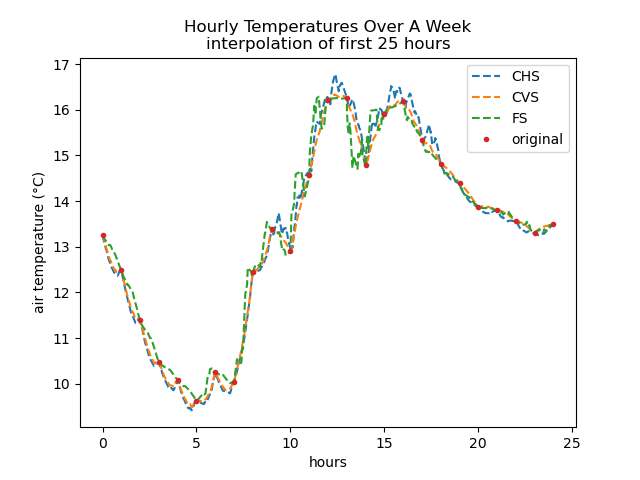}
        \caption{First 25 points}
        \label{fig:Comparison3}
    \endminipage
\end{figure}

We proposed three strategies for the interpolation step, each with a different approach. To obtain a better understanding of the differences between the three methods, we depict in Figures \ref{fig:Comparison1} and \ref{fig:Comparison3} the interpolation results for our Weather data set provided by the three strategies on the same graphic. 
	
We can observe that CHS (with $s_i \in [0,0.2]$) and CVS approaches provide similar results, while the FS approach determines slightly higher variations.
	
To emphasize the comparison, let us use the original Weather data set. We extract hourly data and use the three strategies for fractal interpolation with a number of five interpolation points (\textit{n\_interpolation} = 5) to simulate 10-minute data.
	
\begin{figure}[!htp]
    \minipage{0.33\textwidth}
        \includegraphics[width=\linewidth]{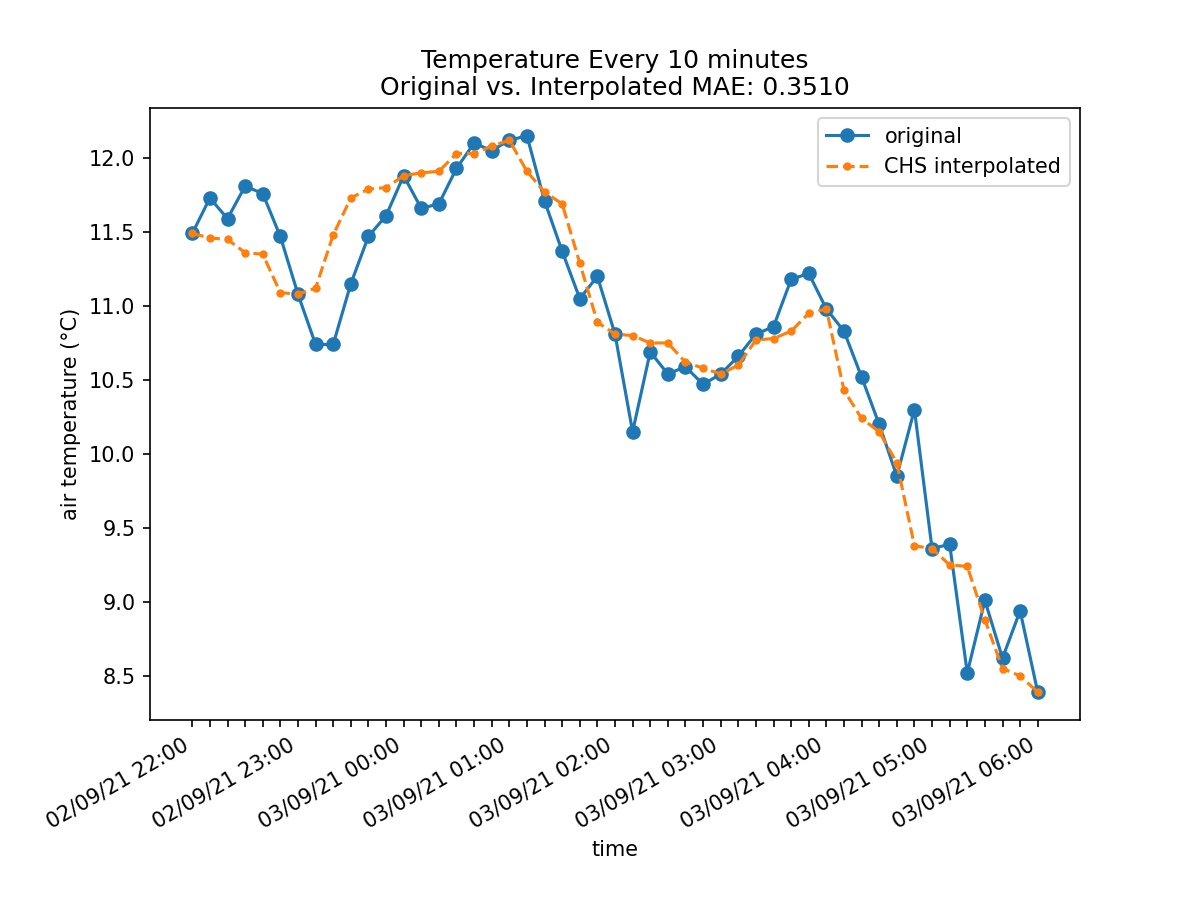}
        \caption{MAE, CHS}
        \label{fig:ComparisonCHS}
    \endminipage
    \hfill
    \minipage{0.33\textwidth}
        \includegraphics[width=\linewidth]{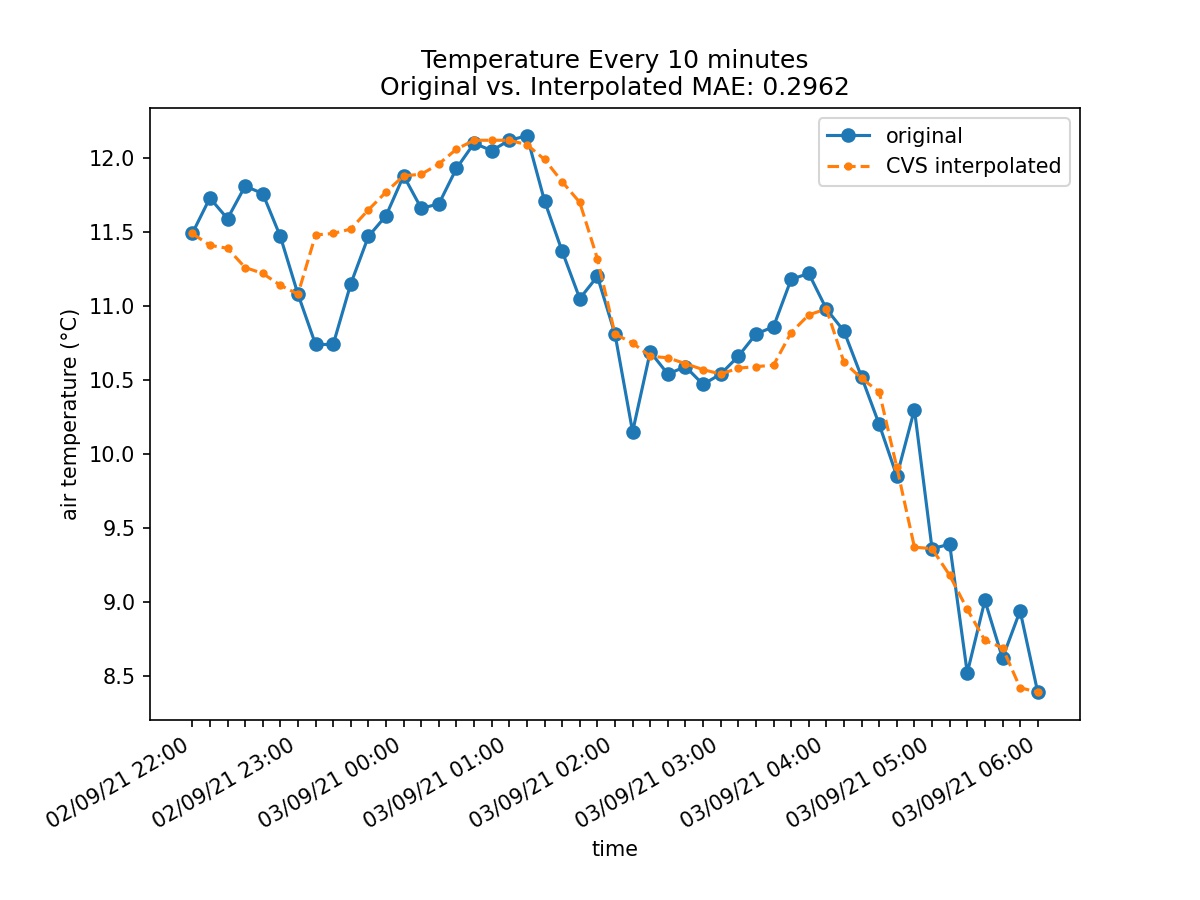}
        \caption{MAE, CVS}
        \label{fig:ComparisonCVS}
    \endminipage
    \minipage{0.33\textwidth}
        \includegraphics[width=\linewidth]{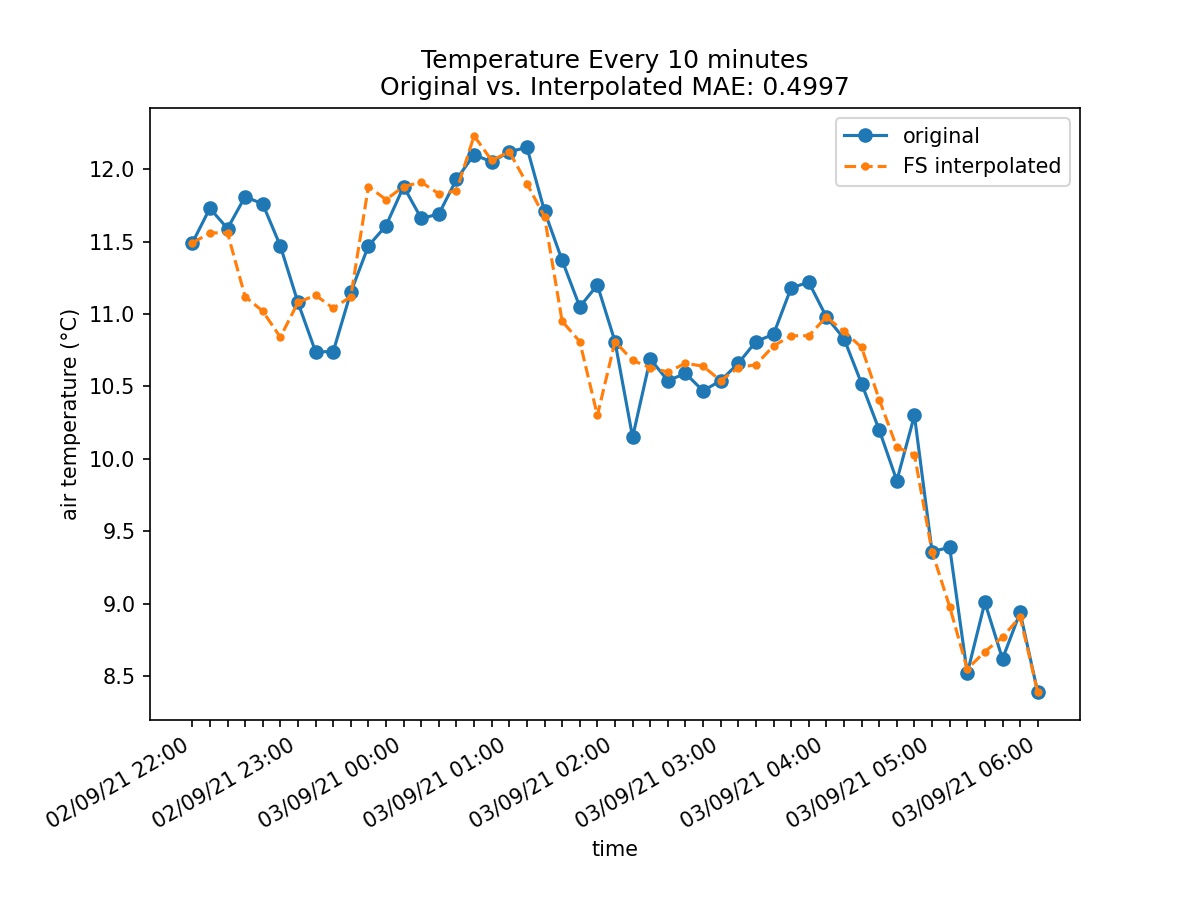}
        \caption{MAE, FS}
        \label{fig:ComparisonFS}
    \endminipage
\end{figure}

In Figures \ref{fig:ComparisonCHS} - \ref{fig:ComparisonFS} there are presented the results for data recorded between  02/09/21, 22:00 and 03/09/21, 06:00 compared to the original data for all three strategies. To obtain a better understanding of the differences between the three strategies, we computed the Mean Absolute Error (MAE) for each data set, which provides us with the mean difference, in degrees, between the real and the interpolated data. 
	
As regards the Weather data set, for CHS we obtain $MAE = 0.3510$, for CVS we have $MAE = 0.2962$ and for FS we get $MAE = 0.4997$. Thus, we can observe that the least MAE is obtained for CVS, thus, this strategy is the optimal one, followed by CHS and FS.
		
\subsubsection{Normalization Step}\label{Normalization Step}
	
Usually, the normalization step's objective is to convert an attribute's values to a better range. There are more strategies to normalize data. We chose to scale the data using a method similar to the \textit{min-max} method, which performs a linear transform of the data in a given interval.
	
More precisely, to normalize data $x'\in[0,1]$ we use the formula
\begin{equation*}
    x'= \dfrac{x- \min x}{\max x -\min x}.
\end{equation*}
For data $x''\in[-1,1]$ the formula becomes
\begin{equation}
    x''= 2\dfrac{x- \min x}{\max x -\min x}-1.
\end{equation}
In general, for data $x'''\in[a,b]$, where $a,b \in \mathbb{R}$ the formula becomes
\begin{equation*}
    x'''= (b-a)\dfrac{x- \min x}{\max x -\min x}+a.
\end{equation*}
	
The purpose of normalization is to transform data in a way that they are either dimensionless and/or have similar distributions. Normalization is an essential step in data preprocessing in any ML algorithm and model fitting. Propagated errors must not have high values, especially in the case of recurrent neural networks (as is LSTM). Moreover, data normalization allows comparison of the results over data with a different configuration, thus reducing biased prioritization of some features over others, which can be caused by providing data with different features that have wide-scale differences to the model.
	
\subsubsection{Stationarity in Time Series Analysis}\label{Stationarity in Time Series Analysis}
	
Stationarity is an important concept in the field of time series analysis with tremendous influence on how the data is perceived and predicted. It indicates whether statistical properties such as mean, variance, and autocorrelation of a time series change over time. 
	
To determine whether a data set must be transformed, we use the Augmented Dicky-Fuller test which uses the coefficient that defines the unit root, p-value. If the p-value obtained is below $0.05$, then the current data set is stationary.
	
For maintaining data stationarity, several transformations can be applied to eliminate trends and seasonality of a  data set. In Table \ref{table1} there are presented the p-values obtained after the three types of transformations used for eliminating the trend: Log Transformation, Square Root Transformation and Linear Regression Transformation.
	
\begin{table}[!htp]
    \centering
    \small
    \begin{tabular}{|l|l|l|l|l|l|}
    \hline
        Data Set &
        \begin{tabular}[c]{@{}l@{}}Interpolation\\ Strategy\end{tabular} &
        None &
        Log &
        Square &
        \begin{tabular}[c]{@{}l@{}}Linear\\ Regression\end{tabular} \\
    \hline
        \multirow{5}{*}{Shampoo Sales}
        & None & 1.0000 & 0.9983 & 0.9991 & \textbf{0.0000} \\
        & Linear & 0.7860 & 0.8076 & 0.7829 & 0.0655 \\
        & CHS & 0.9655 & 0.8245 & 0.9221 & \textbf{0.1022} \\
        & CVS & 0.9755 & 0.9353 & 0.9678 & \textbf{0.0839} \\
        & FS & 0.8744 & 0.0423 & 0.7658 & \textbf{0.0054} \\
    \hline
        \multirow{5}{*}{Air Passengers}
        & None & 0.9919 & \textbf{0.4224} & 0.9181 & 0.4415 \\
        & Linear & 0.0934 & 0.2104 & 0.1522 & 0.0001 \\
        & CHS & 0.1250 & 0.2869 & 0.2118 & \textbf{0.0002} \\
        & CVS & 0.1334 & 0.2248 & 0.1919 & \textbf{0.0001} \\
        & FS & 0.0919 & 0.2018 & 0.1442 & \textbf{0.0001} \\
    \hline
        \multirow{5}{*}{Wheat}
        & None & 0.0607 & \textbf{0.0018} & 0.0122 & 0.9983 \\
        & Linear & 0.4275 & 0.3043 & 0.3704 & 0.8529 \\
        & CHS & 0.4840 & \textbf{0.2979} & 0.3924 & 0.9303 \\
        & CVS & 0.4997 & \textbf{0.3335} & 0.4207 & 0.9110 \\
        & FS & 0.6144 & \textbf{0.4094} & 0.5235 & 0.8590 \\
    \hline
        \multirow{5}{*}{Maize} 
        & None & 0.2370 & \textbf{0.0207} & 0.1073 & 0.9986 \\
        & Linear & 0.4905 & 0.1421 & 0.3132 & 0.9008 \\
        & CHS & 0.4047 & \textbf{0.0497} & 0.1863 & 0.9751 \\
        & CVS & 0.4196 & \textbf{0.0820} & 0.2131 & 0.9798 \\
        & FS & 0.4110 & \textbf{0.0965} & 0.2446 & 0.9493 \\
    \hline
        \multirow{5}{*}{\begin{tabular}[c]{@{}l@{}}Hourly Temperatures\\Over a Week\end{tabular}} 
        & None & \textbf{0.0000} & - & - & - \\
        & Linear & 0.0000 & - & - & - \\
        & CHS & \textbf{0.0000} & - & - & - \\
        & CVS & \textbf{0.0002} & - & - & - \\
        & FS & \textbf{0.0000} & - & - & - \\
    \hline
        \multirow{5}{*}{\begin{tabular}[c]{@{}l@{}}Max Daily \\ Temperature\end{tabular}}
        & None & 0.1210 & 0.8297 & \textbf{0.1160} & 0.8669 \\
        & Linear & 0.0343 & 0.0772 & 0.0510 & 0.4679 \\
        & CHS & \textbf{0.1374} & 0.1942 & 0.1504 & 0.7768 \\
        & CVS & 0.2860 & \textbf{0.2697} & 0.3451 & 0.8406 \\
        & FS & 0.0714 & \textbf{0.0453} & 0.0655 & 0.5549 \\
    \hline
    \end{tabular}
    \caption{p-values obtained for the performed transformations}
    \label{table1}
\end{table}
	
Since for the hourly temperatures over a week, the data set is already stationary and no transformation was required as can be seen in Table \ref{table1}, we manually created a data set composed of the daily maximum temperature recorded. 
	
Daily temperature extremes are more informative than the mean value, indicating the range over this time interval, which is highly important for interpreting its decisive influence on various processes. The minimum temperature generally occurs at dawn, at the end of the negative net radiation interval, when atmospheric status in the boundary layer is rather stable. The situation is completely different as concerns the maximum daily temperature, which arises at midday, after the heating peak, which also induces a considerable enhancement of thermal turbulence, causing increased air temperature variability. Consequently, the daily maximum temperature was chosen for this study, considering that fractal interpolation and a machine learning approach could significantly improve its assessment.

As a result of the different dimensions of the initial and interpolated data set, the p-values obtained are different. For comparison, we also perform linear interpolation. We can notice from Table \ref{table1} that regardless of the interpolation strategy employed, the p-values are close to the value obtained for linear interpolation.
	
We highlight in Table \ref{table1} the transforms that are maintained for each strategy. 
	
As seasonality is regarded, a frequent step is the differentiation of data. However, since fractal interpolation produces functions that are continuous, but are not necessarily everywhere differentiable, this step can not be considered.
	
Moreover, the LSTM model has a sufficient complexity to process non-stationary data, as is not the case for other statistic models like Autoregressive Integrated Moving Average (ARIMA) which depends closely on data stationarity. For the LSTM model, data normalization is more significant. 
	
\subsection{Data Splitting Step}\label{Data Splitting Step}
	
\begin{table}[!htp]
    \centering
    \small
    \begin{tabular}{|c|c|c|}
        \hline
        Data Set & Train Data & Test Data\\
        \hline
        Hourly  & 01/09/21, 00:00 -  & 05/09/21, 21:00 - \\
        Temperature & 05/09/21, 21:00 & 08/09/21, 00:00\\
        \hline
        Daily Maximum Temperature & 01/09/21 - 02/11/21 & 02/11/21 - 30/11/21\\
        \hline
    \end{tabular}
    \caption{Splitting of the Weather data set}
    \label{divisionWeather}
\end{table}
	
For all datasets, 70\% of them are retained, while the remaining 30\% are allocated to the test data set, in chronological order.
	
For the Weather data set, the division of the data set is presented in Table \ref{divisionWeather}.
	
The current division presented in Table \ref{divisionWeather} is relevant for the Weather data set and does not alter the predictions as in the Brasov mountain region, also including the neighbouring depressionary area, the three autumn months (September, October, November) are generally characterized by a more stable weather regime as compared with the symmetrical springtime interval. There are some sudden cold intervals, especially at the end of September or the beginning of October, but after this more dynamic period, the weather pattern shows prolonged intervals of fine weather, favourable for mountain tourism, which is also encountered towards the end of November in many years. 
	
\subsection{Model Description}\label{Model Description}
	
The selected prediction model uses an LSTM layer, followed by one dense layer with dimension $1$ since the datasets have only one feature.
	
\begin{figure}[!htp]
    \centering	
    \vspace{-0.25cm}						 
    \includegraphics[width=0.7\textwidth]{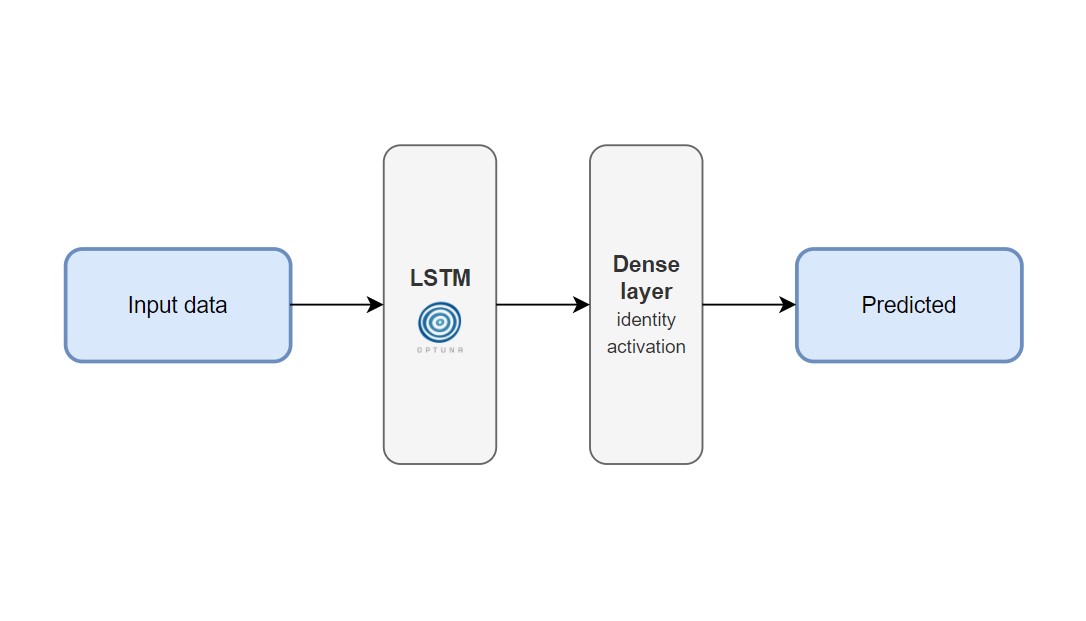}
    \caption{Neural Network Structure}
    \label{fig:model}
    \vspace{-0.25cm}
\end{figure}
	
For each data set, the number of hidden layers in LSTM was optimized using \textit{Optuna}. For the Weather data set, the obtained values are presented in Table \ref{hiddenlayers}.
	
\begin{table}[!htp]
    \centering
    \small
    \begin{tabular}{|c|c|}
        \hline
        Interpolation Strategy & Hidden Layers \\
        \hline
        None & 48 \\
        \hline
        CHS & 39 \\
        \hline
        CVS & 28 \\
        \hline
        FS & 10 \\
        \hline
    \end{tabular}
    \caption{Number of hidden layers}
    \label{hiddenlayers}
\end{table}
	
LSTM is configured with the default parameters. The metric used for measuring the loss in the training step is Mean Squared Error (MSE) and for the general evaluation of the model we used RMSE.
	
\subsubsection{Specific Input Structure of LSTM}\label{Specific Input Structure of LSTM}
	
To be able to feed the data to the LSTM layer, we must transform the data set into a supervised learning format. This is achieved by sliding a window of size \textit{input\_data\_points} over the whole data set using a step of 1. The obtained subsets will represent the inputs for the network. In the case of Univariate Time Series prediction, the output for subset $i$  $[data_i, data_{\text{i+input\_data\_points-1}}]$ will be $data_{\text{i+input\_data\_points}}$. 

The LSTM layer implementation in \textit{Keras - Tensorflow Keras 2.4.1}, expects the input data to be in the format of [\textit{batch\_size}, \textit{input\_data\_points}, \textit{features}]. In our case, we used a batch size of 1.
	
	Special attention is needed such that the window size  \textit{input\_data\_points} considered must be less than 30\% of the entire dataset. Otherwise, using the testing set to evaluate the model's performance will not be possible, as for a single data point prediction, \textit{input\_data\_points} entries are required in the supervised format.

\subsubsection{Model Optimization}\label{Model Optimization}
	
Optimization plays an important role in a machine learning solution and the step of tuning a chosen model is critical to the model’s performance and accuracy. Hyperparameter tuning intends to find the hyperparameters of a given machine learning algorithm that guarantee the best performance as measured on a validation set.  Saving time, but also eliminating the chance of overfitting or underfitting influenced the interest in hyperparameter tuning research. Thus, entire branches of machine learning and deep learning theory have been dedicated to the optimization of models.
	
In our study, we use \textit{Optuna - Optuna 2.10.0} in two key points of our implementation:

\begin{itemize}
    \item[1.]	optimizing the vertical scaling factor $s_i$ for the interpolation step;
    \item[2.]	fine-tuning LSTM network hyperparameters.
\end{itemize}
	
For the LSTM network model, we considered the following hyperparameters for optimization:

\begin{itemize}
    \item[$\circ$] \textit{units}: the number of hidden layers used for the LSTM layer, which defines the complexity of the model. There is no formula or rule to determine this number, so it is a natural decision to pick this hyperparameter for the optimization process
    search space. Our choice is the interval $[2, 64]$;
    
    \item[$\circ$] \textit{input\_data\_points} (timesteps): given the data structure expected by the LSTM network, this represents the number of consecutive input values considered for an output value in the supervised learning format. Since this is highly dependent on the data set size, but also on the frequency of the desired prediction, we considered the interval $[1, min(x,30 * \text{length(dataset)}/100 - 1)]$, where $x = 15$ for non-interpolated datasets and $x=100$ for the interpolated ones. This limit was necessary for the optimization process because a large window size produces fewer predictions which misleadingly leads to smaller cumulated errors, however, the purpose of the model is to come up with predictions relying on as few known values as possible.
    
    \item[$\circ$] \textit{learning\_rate}: finding the right value for the learning rate can significantly improve the accuracy of the model. A learning rate too big might lead to poor performance since the algorithm makes leaps too large while searching for the optimal value, whereas a small learning rate slows down the execution time of the training process. For the search space, we chose the interval $[1e-3, 1e-1]$.
    
    \item[$\circ$]\textit{epochs}: although not specific to Recurrent neural networks (RNNs), the number of epochs to train a model is an important key factor. The more epochs, the better for the model, but too many epochs can often lead to overfitting. Each model corresponding to a non-interpolated data set was trained for 150 epochs, while the interpolated ones were limited to 25. The final epoch number was hand-picked by observing the evolution of the loss at each epoch
\end{itemize}	
	
As described in Section \ref{Optuna Framework}, the \textit{suggest} APIs from \textit{Optuna} framework are used to define the search space for each selected hyperparameter. Each study ran 50 trials since no significant improvement in the overall score was observed after that.
	
The objective function defined will create, compile, fit and evaluate the model using the training data set 5 times for each set of considered hyperparameters. This is done to ensure that the optimized hyperparameters produce a more stable model. After the optimization, we obtained the configurations from Table \ref{table2} with the corresponding scores. Mean results are again computed for 5 individual runs using the same configuration.
	
\begin{table}[]
    \centering
    \resizebox{\textwidth}{!}{%
        \begin{tabular}{|l|l|l|l|l|l|l|l|}
            \hline
                Data Set & 
                \begin{tabular}[c]{@{}l@{}}Linear\\ Regression\end{tabular} & 
                \begin{tabular}[c]{@{}l@{}}Hidden\\ Layers\end{tabular} &
                \begin{tabular}[c]{@{}l@{}}Input Data\\ Points\end{tabular} &
                Epochs &
                \begin{tabular}[c]{@{}l@{}}Learning\\ Rate\end{tabular} &
                \begin{tabular}[c]{@{}l@{}}Train\\ RMSE\end{tabular} &
                \begin{tabular}[c]{@{}l@{}}Test\\ RMSE\end{tabular} \\
            \hline
                \multirow{4}{*}{\begin{tabular}[c]{@{}l@{}}Shampoo\\ Sales\end{tabular}}
                & None & 62 & 8 & 80 & 0.02 & 0.1717 & 0.4997 \\
                & CHS & 48 & 16 & 15 & 0.04 & 0.0859 & 0.0951 \\
                & CVS & 41 & 28 & 10 & 0.01 & 0.0607 & \textbf{0.0749} \\
                & FS & 17 & 1 & 8 & 0.005 & 0.1218 & 0.1264 \\
            \hline
                \multirow{4}{*}{\begin{tabular}[c]{@{}l@{}}Air\\ Passengers\end{tabular}} 
                & None & 56 & 14 & 65 & 0.007 & 0.0947 & 0.1348 \\
                & CHS & 33 & 97 & 8 & 0.03 & 0.0285 & 0.0655 \\
                & CVS & 62 & 51 & 5 & 0.02 & 0.0288 & \textbf{0.0444} \\
                & FS & 29 & 93 & 7 & 0.03 & 0.0416 & 0.0619 \\
            \hline
                \multirow{4}{*}{Wheat} 
                & None & 13 & 5 & 125 & 0.005 & 0.1480 & 0.2410 \\
                & CHS & 8 & 93 & 12 & 0.03 & 0.0559 & 0.0653 \\
                & CVS & 11 & 98 & 10 & 0.01 & 0.0431 & \textbf{0.0552} \\
                & FS & 51 & 99 & 12 & 0.01 & 0.0656 & 0.0932 \\
            \hline
                \multirow{4}{*}{Maize} 
                & None & 52 & 3 & 150 & 0.05 & 0.1327 & 0.1766 \\
                & CHS & 32 & 82 & 15 & 0.01 & 0.0299 & 0.0304 \\
                & CVS & 39 & 1 & 10 & 0.002 & 0.0187 & \textbf{0.0273} \\
                & FS & 27 & 20 & 20 & 0.007 & 0.0920 & 0.0991 \\
            \hline
                \multirow{4}{*}{\begin{tabular}[c]{@{}l@{}}Hourly \\ Temperatures \\ Over A Week\end{tabular}} 
                & None & 48 & 8 & 60 & 0.008 & 0.1111 & 0.1383 \\
                & CHS & 39 & 22 & 7 & 0.005 & 0.0325 & 0.0414 \\
                & CVS & 28 & 34 & 15 & 0.001 & 0.3446 & \textbf{0.0389} \\
                & FS & 10 & 17 & 14 & 0.005 & 0.0391 & 0.0462 \\
            \hline
                \multirow{4}{*}{\begin{tabular}[c]{@{}l@{}}Daily Max \\ Temperatures\end{tabular}} 
                & None & 43 & 5 & 175 & 0.02 & 0.1132 & 0.4944 \\
                & CHS & 47 & 84 & 12 & 0.03 & 0.0321 & \textbf{0.0523} \\
                & CVS & 49 & 11 & 20 & 0.005 & 0.0290 & 0.0692 \\
                & FS & 50 & 27 & 15 & 0.005 & 0.0526 & 0.1351 \\
            \hline
        \end{tabular}%
    }
    \caption{Model Tuning and Hyperparameters Optimization}
    \label{table2}
\end{table}
	
We mention the fact that all optimizations were executed using the free environment offered by Google Colab, having a GPU-accelerated runtime, making this accessible to everybody.
	
We notice that for the majority of the datasets, the best results are obtained via CVS. However, the other two strategies are also of considerable importance. For all datasets studied, the values obtained for CHS and CVS are rather close, while the results obtained for FS are higher. Even so, FS is a viable strategy for its easy way of usage. Moreover, it is worth noticing that all results obtained with the three strategies are at least 50\% better than the results obtained without any interpolation strategy.
	
\section{Results and Discussions}\label{Results and Discussions}
	
We present the graphics of the predictions obtained via the LSTM model for our Weather data set, for the three proposed strategies and the initial data which is not interpolated. The results emphasize the advantages brought to the predictions. 
	
\begin{figure}[!htp]
    \minipage{0.45\textwidth}
        \includegraphics[width=\linewidth]{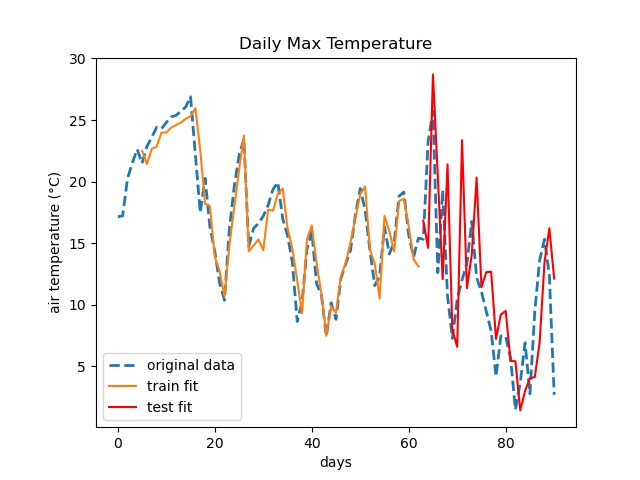}
        \caption{Prediction for daily maximum data without interpolation}
        \label{fig:maxNone}
    \endminipage
    \hfill
    \minipage{0.45\textwidth}
        \includegraphics[width=\linewidth]{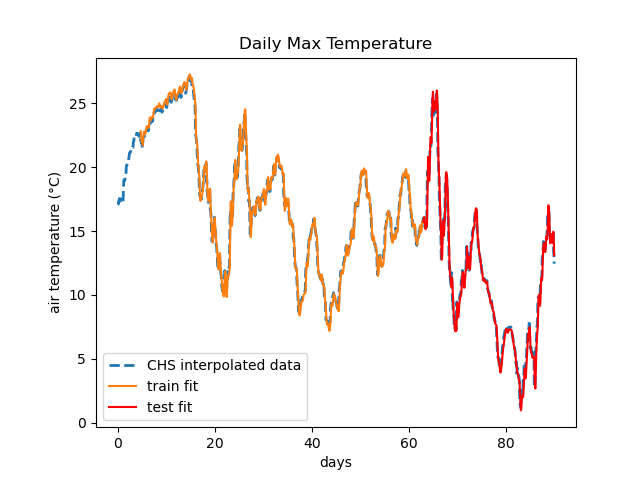}
        \caption{Prediction for daily maximum data with CHS}
        \label{fig:maxCHS}
    \endminipage
\end{figure}

\begin{figure}[!htp]
    \minipage{0.45\textwidth}
        \includegraphics[width=\linewidth]{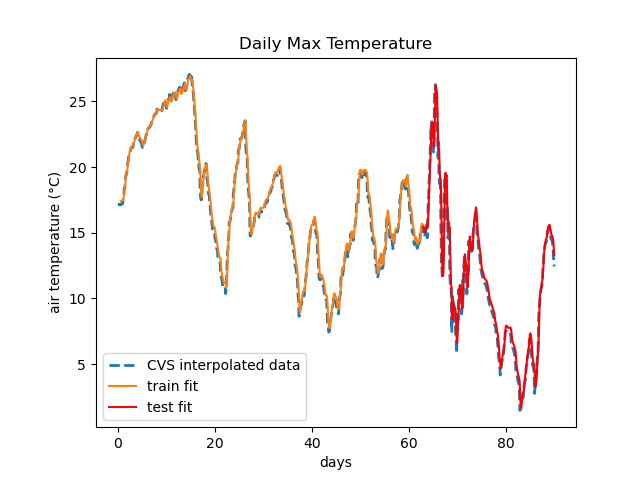}
        \caption{Prediction for daily maximum data with CVS}
        \label{fig:maxCVS}
    \endminipage
    \hfill
    \minipage{0.45\textwidth}
        \includegraphics[width=\linewidth]{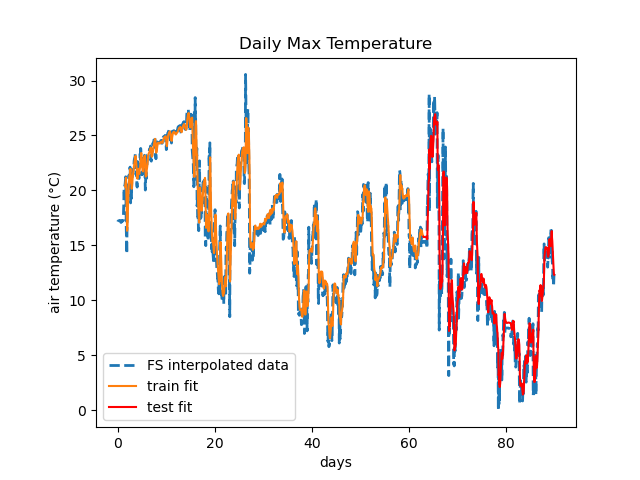}
        \caption{Prediction for daily maximum data with FS}
        \label{fig:maxFS}
    \endminipage
\end{figure}

For the manufactured data set which contains the daily maximum temperatures, the results are presented in Figures \ref{fig:maxCHS}-\ref{fig:maxFS}. It can be observed that for the initial data, the LSTM does not perform well enough. However, for the interpolated datasets, the results are visibly better, with the best result being obtained for CHS. All the strategies provide better results and this is motivated by the fact that more training data provides better ML solutions. Thus, the importance of the proposed preprocessing strategies is supported and emphasized. 
	
\begin{figure}[!htp]
    \minipage{0.45\textwidth}
        \includegraphics[width=\linewidth]{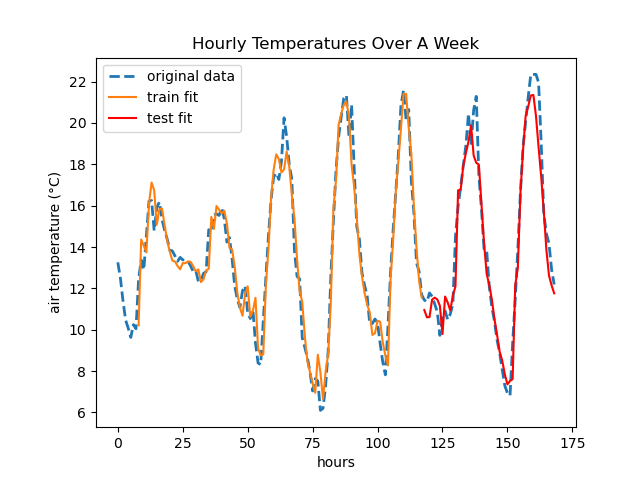}
        \caption{Prediction for hourly data without interpolation}
        \label{fig:HourlyNone}
    \endminipage
    \hfill
    \minipage{0.45\textwidth}
        \includegraphics[width=\linewidth]{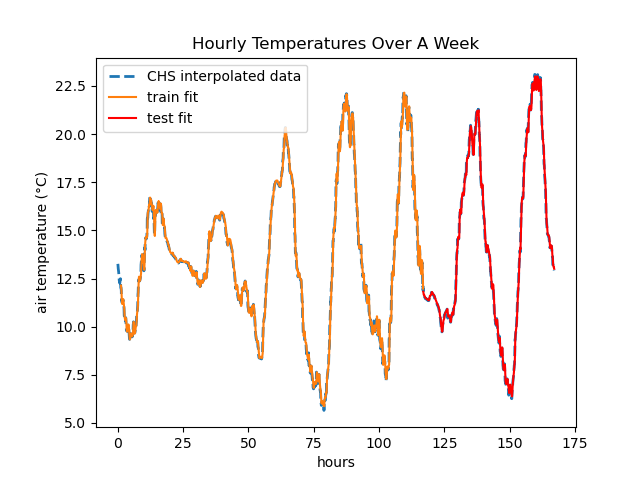}
        \caption{Prediction for hourly data without CHS}
        \label{fig:HourlyCHS}
    \endminipage
\end{figure}
	
\begin{figure}[!htp]
    \minipage{0.45\textwidth}
        \includegraphics[width=\linewidth]{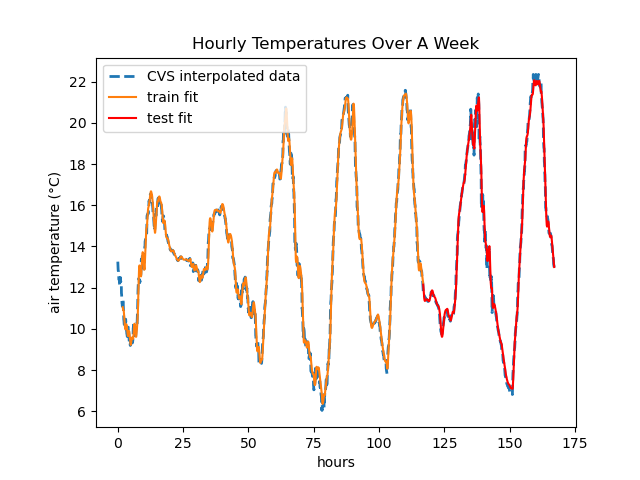}
        \caption{Prediction for hourly data with CVS}
        \label{fig:HourlyCVS}
    \endminipage
    \hfill
    \minipage{0.45\textwidth}
        \includegraphics[width=\linewidth]{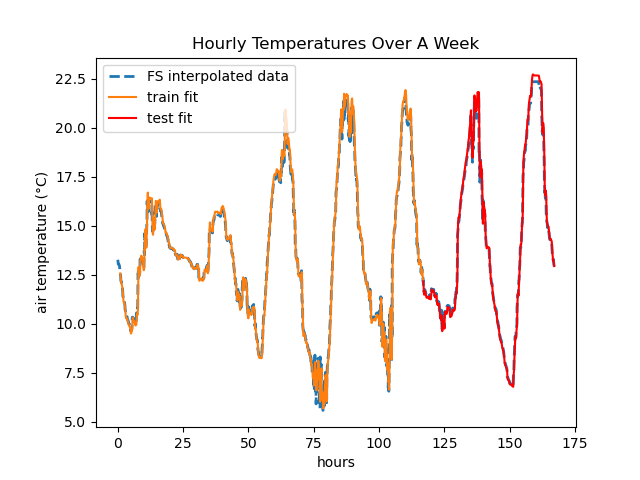}
        \caption{Prediction for hourly data with FS}
        \label{fig:HourlyFS}
    \endminipage
\end{figure}

The results for both the maximum daily temperature (considered for example purposes) and the hourly data set with 17 interpolation points, see Figures \ref{fig:HourlyCHS} - \ref{fig:HourlyFS}, prove that fractal interpolation brings a significant upgrade to the model as it improves visibly the results of the predictions when compared to the results obtained for the initial datasets (results that are presented in Figure \ref{fig:maxNone} and \ref{fig:HourlyNone}). Even though 17 interpolation points provide good theoretical results, we should also practically test the predictions by considering the hourly data set with 5 interpolation points to simulate the 10-minute data recorded by the sensor.
	
Thus, we consider the entries from the Weather data set from 01/09/21, 00:00 to 08/09/21, 00:00. For this data set, we extract the hourly temperatures data and use fractal interpolation with \textit{n\_interpolation} = 5 according to the three strategies proposed (CHS, CVS, FS) to simulate 10-minute data.

\begin{table}[]
    \centering
    \resizebox{\textwidth}{!}{%
        \begin{tabular}{|l|l|l|l|l|l|l|l|}
            \hline
                Data Set & 
                \begin{tabular}[c]{@{}l@{}}Linear\\ Regression\end{tabular} & 
                \begin{tabular}[c]{@{}l@{}}Hidden\\ Layers\end{tabular} &
                \begin{tabular}[c]{@{}l@{}}Input Data\\ Points\end{tabular} &
                Epochs &
                \begin{tabular}[c]{@{}l@{}}Learning\\ Rate\end{tabular} &
                \begin{tabular}[c]{@{}l@{}}Train\\ RMSE\end{tabular} &
                \begin{tabular}[c]{@{}l@{}}Test\\ RMSE\end{tabular} \\
            \hline
                \multirow{4}{*}{\begin{tabular}[c]{@{}l@{}}Hourly \\ Temperatures \\ Over A Week\end{tabular}} 
                & None & 48 & 8 & 60 & 0.008 & 0.1111 & 0.1383 \\
                & CHS & 30 & 94 & 15 & 0.03 & 0.0415 & \textbf{0.0462} \\
                & CVS & 61 & 94 & 10 & 0.02 & 0.0443 & 0.0474 \\
                & FS & 34 & 99 & 17 & 0.016 & 0.0488 & 0.0494 \\
            \hline
                \begin{tabular}[c]{@{}l@{}}Temperature\\ Every 10 minutes\end{tabular} 
                & None & 55 & 32 & 25 & 0.016 & 0.0523 & 0.0541 \\
            \hline
        \end{tabular}%
    }
    \caption{Comparison between interpolated 10-minute data and original data predictions}
    \label{tab10min}
\end{table}
	
We can observe from Table \ref{tab10min} that while the hourly data without any interpolation provides visibly worse results (as expected) than the predictions with initial 10-minute data, the case is the opposite when interpolating the hourly data set to simulate 10-minute data with either of the three strategies. Thus, it can be observed that extracting hourly data and constructing artificial fractal interpolation points for 10-minute data is a better strategy for prediction. The result is surprising and requires further testing on various weather datasets, but the current results obtained for the considered data set and the consistent differences between the test RMSE results, see Table \ref{tab10min}, allow us to be optimistic about the performances of our strategies on real-world data (especially meteorological datasets).
	
\begin{figure}[!htp]
    \minipage{0.45\textwidth}
        \includegraphics[width=\linewidth]{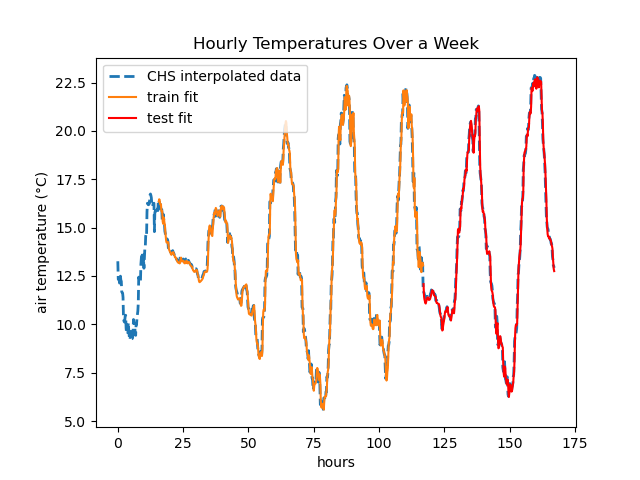}
        \caption{Prediction for hourly data set interpolated with CHS, n\_interpolation=5}
        \label{fig:Hourly10Min}
    \endminipage
    \hfill
    \minipage{0.45\textwidth}
        \includegraphics[width=\linewidth]{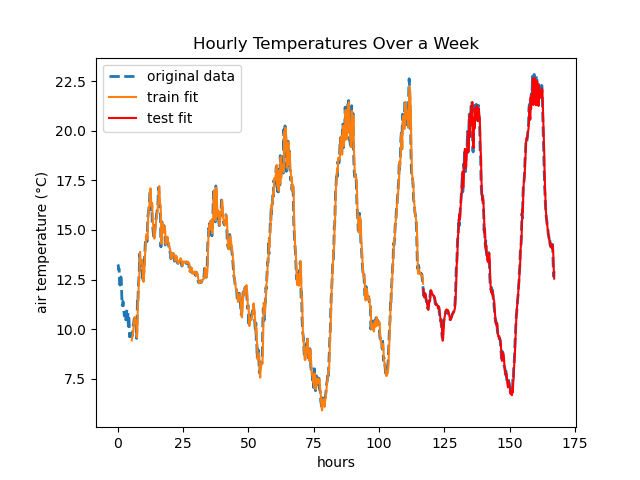}
        \caption{Prediction using the original 10-minute entry data set}
        \label{fig:Hourly10MinNone}
    \endminipage
\end{figure}
	
Artificial Neural Networks (ANN) and machine learning have proven to be efficient tools for predicting meteorological data. These could be significant for predicting data for smaller local areas since some meteorological processes are too small-scale or too complex to be explicitly included in classical numerical weather prediction models. 
	
Thus, our study proposes a recent idea of using fractal interpolation tools for preprocessing data before feeding the data to an ML algorithm, in our case, an LSTM algorithm.

 Without any doubt, this approach has some limitations, and there are opportunities to improve this study. 
    
In order to provide a comprehensive coverage of the interpolation step, the study proposes three techniques. The persistence of certain aspects of the data is not guaranteed by the stop condition from the CHS method, although it assures that the Hurst exponent for the interpolated data is sufficiently near to the original Hurst value. This inspired us to develop novel interpolation techniques guaranteeing the preservation of specific data features. The final approach, namely FS, similarly focuses on maintaining these characteristics while also improving in terms of time complexity.
    
Furthermore, future experiments could be completed in order to assess the performance of the proposed strategies in relation to outliers, but also further testing on various weather datasets from different regions needs to be performed. In addition, although the LSTM modelling obtained is sufficient for the selected datasets, a more complex model can be developed to achieve better accuracy results.

\section{Conclusions}\label{Conclusions}

The results obtained in this study confirm the relevance and extend the applications of fractal interpolation as a time series augmentation technique. The three proposed strategies involve optimizing the vertical scaling factor and the size of the fractal interpolation subset to generate relevant data in the context of the prediction optimization problem. Depending on the source domain and data pattern, we were able to identify the appropriate interpolation strategy in order to improve predictions. As a result, for all considered datasets, the current approach improved accuracy prediction results between 50\% and 89\% over the base case where the raw data was used.

By using the meteorological dataset of temperatures recorded in Bra\c sov, we were able to show that the three proposed strategies can also be used independently of a prediction model in order to obtain data simulating a higher sampling rate than the maximum capacity of the sensor, with an average error of maximum $\pm 0.49$. Numerical weather prediction models such as those described in \cite{Malardel}, based on fluid dynamics and thermodynamic equations, could also benefit from this data enrichment. Although they do not outperform solutions based on artificial neural networks, these models can perform well for short time intervals (up to 5 days) when sufficient data are available.

We have highlighted the need to use machine learning modelling in this study. In addition, it is possible to go even further with the results obtained so that a relevant prediction on meteorological data also implies a focus on process optimization in precision agriculture, early detection of extreme weather phenomena, and local and global environmental understanding.

Our outcomes extend the current results existing in the literature and contribute significantly to research dedicated to data augmentation and data preprocessing, as well as enhancing machine learning prediction models. Moreover, our results provide a significant answer to the question of refining data prediction based on data recorded at larger intervals.

\vspace{0.5cm}

\textbf{Data Availability Statement} The datasets generated during and/or analysed during the current study are available from the corresponding author upon reasonable request.
	
\textbf{Funding and Conflicts of interests/Competing interests} The authors declare no conflict of interest. Funding information is not applicable/No funding was received.
 
\bibliographystyle{elsarticle-harv} 
\bibliography{main.bib}

\end{document}